\documentclass[journal]{IEEEtran}
\usepackage{graphicx}
\ifCLASSINFOpdf
\else
\fi
\usepackage{algorithmic}

\usepackage{amsmath}
\usepackage{hyperref}
\usepackage{algorithmic}
\usepackage{subfigure}
\usepackage[linesnumbered,ruled,vlined]{algorithm2e}
\SetKwInput{KwInput}{Input}                
\SetKwInput{KwOutput}{Output}              

\begin{document}
%
\title{Unsupervised Flood Detection on SAR Time Series}
%
%
%

\author{Ritu Yadav\thanks{This research is part of the project ‘EO-AI4GlobalChange’ funded by Digital Future.}, Andrea Nascetti, Hossein Azizpour, Yifang Ban,~\IEEEmembership{Member,~IEEE,}

\thanks{Ritu Yadav, Andrea Nascetti and Yifang Ban is with Division of Geoinformatics, KTH Royal Institute of Technology, Sweden. (e-mail: rituy@kth.se, nascetti@kth.se, yifang@kth.se)}
\thanks{Hossein Azizpour is with Department of Robotics, Perception and Learning (RPL), KTH Royal Institute of Technology, Sweden. (e-mail: azizpour@kth.se)}
}

\maketitle

\begin{abstract}
Human civilization has an increasingly powerful influence on the earth system. Affected by climate change and land-use change, natural disasters such as flooding have been increasing in recent years. Earth observations are an invaluable source for assessing and mitigating negative impacts. Detecting changes from Earth observation data is one way to monitor the possible impact. Effective and reliable Change Detection (CD) methods can help in identifying the risk of disaster events at an early stage. 
In this work, we propose a novel unsupervised CD method on time series Synthetic Aperture Radar~(SAR) data. Our proposed method is a probabilistic model trained with unsupervised learning techniques, reconstruction, and contrastive learning. The change map is generated with the help of the distribution difference between pre-incident and post-incident data. Our proposed CD model is evaluated for flood detection task. We verified the efficacy of our model on 8 different flood sites, including three recent flood events from Copernicus Emergency Management Services and six from the Sen1Floods11 dataset.
Our proposed model achieved an average of 64.53\% Intersection Over Union(IoU) value and 75.43\% F1 score. Our achieved IoU score is approximately 6-27\% and F1 score is approximately 7-22\% better than the compared unsupervised and supervised existing CD methods. Based on our CD method, we also proposed an automatic change point detection framework where time series data is processed through the model to identify percentage change and the date on which significant change started to reflect on SAR data. This can help in early detection of floods giving more time for response. Our proposed model and change point detection framework are lightweight and easy to deploy. We conducted a range of experiments and ablation on our model. The results and extensive discussion presented in the study show the effectiveness of the proposed unsupervised CD method.
\end{abstract}


\IEEEpeerreviewmaketitle

\section{Introduction}
\label{sec:intro}
\IEEEPARstart{A}{ccording} to a report from the Centre for Research on the Epidemiology of Disasters (CRED) \cite{disaster_report}, In 2021, a total of 432 catastrophic events were recorded, which is considerably higher than the average of 357 annual catastrophic events for 2001-2020. Floods dominated these events, with 223 occurrences, up from an average of 163 annual flood occurrences recorded across the 2001-2020 period. Countries such as India, China, Afghanistan, and Germany faced the loss of thousands of lives and billions of dollars~\cite{disaster_report}.
Current flood predictions and evacuation services are gradually improving and are not fully reliable to handle the situation before a flood. In most flooding events, rescue services are launched afterward. In such a scenario, accurate and reliable flood maps reflecting the damaged areas can help in efficient emergency response. The maps can be used for rescue missions, re-routing traffic, delivering aid, and many more.
In the case of large-scale floods, on-ground evaluation for the identification of affected areas can be risky due to unfavorable weather conditions and collapsed transportation systems. Whereas, satellites can help in quick access to ground information over a large geographical area. The data can be used in detecting and mapping flooded areas and their severity.
Satellites are a leading technology in gathering quick information on a large scale. A rapid increase in remote sensing technology leads to an immense amount of earth observation sensors providing data at different spectral, spatial, and temporal resolutions. Compared to optical data, Synthetic Aperture Radar (SAR) imagery is preferred for flood mapping~\cite{anusha2020flood}. Unlike optical sensors, SAR has the capability of imaging day and night, irrespective of the weather conditions. 

Water surface can be detected using SAR because the water surface is smooth and SAR backscatter from a smooth surface is very low \cite{di2011timely}. As a result, the water surface appears in a darker tone whereas the land surface with rough soil texture, building, vegetation, and others appears in bright tones. During floods, land surfaces are partially covered with water causing a significant change in backscatter. Therefore, floods can be detected with a CD framework that is proficient in detecting these backscatter changes.

\section{Related Work}
\label{sec:related_work}

In recent years, deep learning in Earth observation has received significant attention. 
Before deep learning there were classical unsupervised CD methods such as ImageDiff which is simply the difference between bi-temporal images, ImageRatio \cite{lu2004change} uses the ratio of two bands, ImageRegr \cite{luppino2019unsupervised}, CVA \cite{malila1980change} which is a conceptual extension of image differencing. Several superpixels and spatial neighborhood-based variants of CVA have been proposed, such as parcel change vector analysis (PCVA) \cite{bovolo2008multilevel} and robust change vector analysis (RCVA) \cite{thonfeld2016robust}. DPCA \cite{deng2008pca}, and PCDA \cite{dharani2021land} are examples of principal component analysis used for land cover CD. 

Although there exist several classical methods to detect changes in multi-temporal images, deep learning gained new achievements due to its powerful discriminative ability \cite{asokan2019change}.
FC-EF \cite{daudt2018fully} and FC-Siam-diff \cite{bandara2022revisiting, daudt2018fully} are fully connected siamese network variations developed for CD. DS-IFN \cite{zhang2020deeply} proposed a fusion network for bitemporal CD on high-resolution optical images. There are multiple works on attentive siamese networks, DASNet \cite{chen2020dasnet} proposed a change map using L2 distance between the attentive feature maps from a dual siamese encoder, ADS-Net \cite{wang2021ads} proposed a multiscale siamese encoder followed by an attentive decoder and DAUSAR \cite{yadav2022attentive} proposed a dual stream attentive U-Net architecture to detect changes. DASNet and ADS-Net operate on optical imagery whereas DAUSAR detects changes in SAR imagery. There are few works of CD with a generative network such as BIT \cite{chen2021remote} proposed a tokenized transformer network embedded with a deep difference-based CD framework. BIT detects changes in bitemporal high-resolution optical imagery. Deep learning CD methods in remote sensing are predominantly supervised. The above-mentioned methods are some of well-known supervised deep learning methods for CD.

Deep neural networks harness their great feature learning power from a large amount of labeled data. Unfortunately, in earth observation labeling data is a time taking task and requires domain expertise. The challenge is further elevated by the low-resolution data causing difficulty in feature discrimination for data labeling. Due to these challenges, there are not many large-scale datasets in earth observation.
Apart from urban/building monitoring and land cover classification other earth observation task such as flooding, landslide, and wildfire suffers severely due to lack of labeled data. Training supervised networks on small datasets raises questions about their generalizability to other sites. In fact, many studies in earth observation are being conducted on a single site \cite{munoz2021local, manjusree2012optimization, zhong2017spectral}. On the other hand, we have a large amount of unlabeled earth observation data which is readily available for use. An unsupervised CD method can be trained on these unlabeled data and can give more generalized results in comparison to supervised methods trained on small labeled datasets \cite{noh2022unsupervised, jing2022remote, ruuvzivcka2021unsupervised}.
More recently, unsupervised deep learning methods are proposed for CD on remote sensing data such as \cite{zhan2018log} where a denoising autoencoder (DAE) is proposed and \cite{liu2016deep} proposed a highly coupled convolutional network for detecting changes between SAR and optical images. These unsupervised methods are tested on scenes with limited spatial complexity and didn't explore time series data. One of the recent work \cite{ruuvzivcka2021unsupervised} trained simple variational autoencoder on reconstruction task and used distance metric on latent parameters to get low resolution change maps. This network is trained on time series data and designed to detect changes between two Sentinel-2 multispectral images. 

In past few years, unsupervised learning techniques like SimCLR~\cite{chen2020simple}, MoCo \cite{chen2020improved}, BYOL~\cite{grill2020bootstrap} and DeepCluster \cite{caron2018deep} has shown tremendous success. SimCLR and MoCo proposed contrastive learning from 'positive pairs'(augmented version of the same image) and 'negative pairs' (augmented version of a different image). These methods need careful treatment for negative pairs by relying on large batches or memory banks. The need for negative pairs was eliminated by BYOL, relying on learning from positive pairs. DeepCluster is a clustering method that jointly learns the parameters of a neural network and cluster assignments of the resulting features. \cite{dong2021multiscale} used deepcluster method and implemented unsupervised clustering with CNN to learn clustering-friendly feature representations of SAR data.
There are multiple CD works on heterogeneous remote sensing data such as \cite{saha2021self} used both deep cluster and contrastive learning to train a multisensor siamese CD network. 
Another recent work is Code Aligned Autoencoder (CAA) \cite{luppino2022code} where an encoder-decoder network learns features from cross modality with the help of contrastive learning. The network generate output which merge features from the two modalities and looks like somewhere in-between the two modalities. The change map is produced by calculating difference image between the pre change input and generated output followed by manual thresholding. One of the recent work RaVAEN \cite{ruuvzivcka2021unsupervised} trained a simple variational autoencoder on reconstruction task and used distance metric on latent parameters to obtain low-resolution change maps. This network is trained on time series data and designed to detect changes between two Sentinel-2 multispectral images. 

Inspired by unsupervised deep learning techniques, we targeted our challenging problem of CD in a fully self-supervised manner. In this study, we introduce a generative network for CD on Sentinel-1 SAR data. We named our CD method as \textbf{C}ontrastive Conv\textbf{L}STM \textbf{V}ariational \textbf{A}uto\textbf{e}ncoder (CLVAE). Our method utilizes unlabeled time series data to train our network without external supervision at any stage. The key contributions of our work are as follows.
\begin{enumerate}
\item We propose a novel self-supervised CD method that gains its ability predominantly from the strong latent representations learned by the probabilistic reconstruction architecture of the variational autoencoder. We acknowledge the ability of time series data in CD task and to accommodate the benefits we embrace our proposed network with convolutional long short-term memory.

\item We show how the learned latent parameters of a reconstruction network can be employed to generate change maps. See subsection \ref{sec:changeDetect} and framework \ref{fig3}. We further empowered our reconstruction network with cross connections between encoder and decoder branches similar to U-net.

\item We present network trained in a fully self-supervised manner. Along with reconstruction loss, the network is trained using contrastive learning where layers can learn to reconstruct SAR input so well that they can differentiate between dissimilar patches. We followed the contrastive learning idea from MoCo and simplified it for our remote sensing CD task. See training pipeline \ref{fig2} and subsection \ref{sec:Objective_function} for explanation.

\item Our training network is lightweight with 577,239 total parameters. The inference network is even smaller as it only uses the encoder part of the trained network. Both training and inference networks are memory efficient, making it easier for testing and deployment.

\item We display adaptability of unsupervised learning on sparse spatiotemporal SAR satellite data that are substantially different from the natural images commonly used in computer vision. 
\end{enumerate}

\begin{figure*}  
\centering  
\begin{subfigure}
  \centering  
  \includegraphics[trim=0cm 0cm 0cm 0cm, width=170mm, height=80mm]{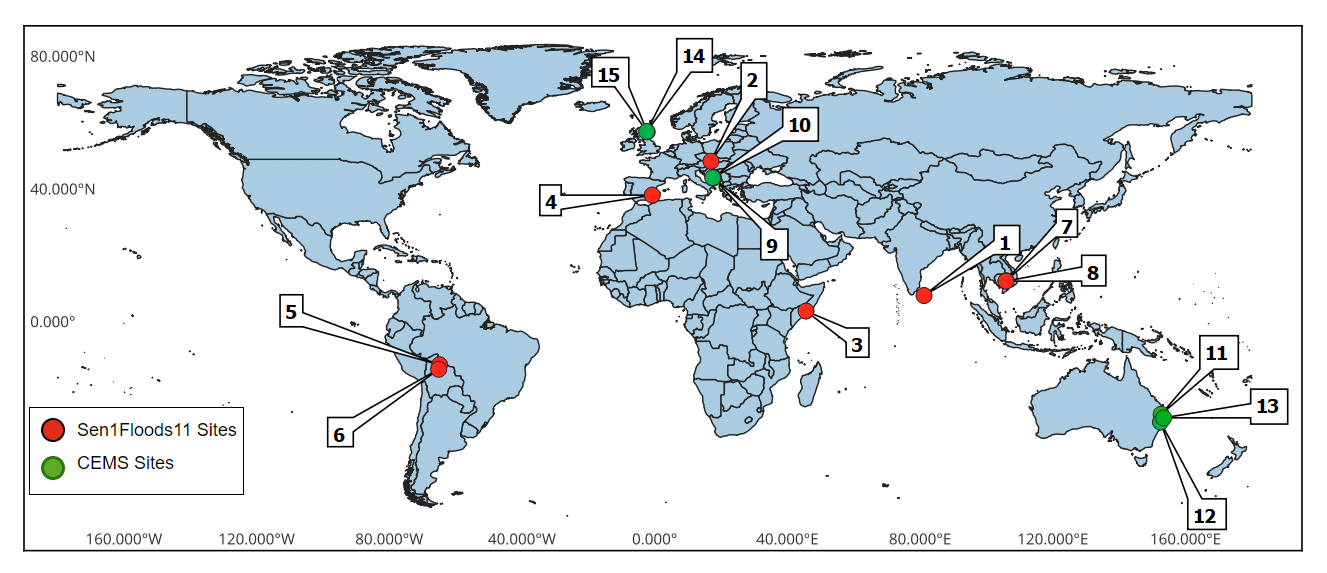}
\end{subfigure}
\caption{Overview of the Study Sites. Colored dots represents tile locations and numbers are the references given to each tile. Red dots represents Sen1Floods11 sites and green dots represents CEMS sites.}
\label{data_study}
\end{figure*}
Additionally, we propose a change point detection framework (see Figure \ref{fig4}). Change point detection aims to locate abrupt property changes in time series data \cite{luppino2022code}. We can use the framework for continuous change monitoring, event detection and temporal anomaly detection such as detecting the point when the change started \cite{deldari2021time}. A significant change is an indicator of a major activity and might require human attention.
\section{Data and Study Area}
\label{sec:Dataset}
Our proposed unsupervised CD method was validated on Sentinel-1 SAR data. Two sources Sen1Floods11~\cite{bonafilia2020sen1floods11} and Copernicus Emergency Management Service (CEMS)\cite{CEMS} were utilised to prepare the data. Collectively, our method was validated on data from 9 flood sites. The location and details of these flood events are presented in Figure~\ref{data_study} and Table~\ref{Flood_data_spec}. The data collection process from both sources is explained in Subsection \ref{subsec:data_collection} followed by the data pre-processing steps in Subsection \ref{subsec:satellite_data_preprocessing}

\begin{table}[htbp]
\caption{Flood Event Metadata. 'Tile Ref.' is the reference number given to each tile, Site is the name of the flood site, 'S1 Post Date' is the date of the Sentinel-1 post-image, 'GT Date' is the date of the satellite image which is used to create the ground truth, 'Rel. Orbit' is the relative orbit number of the Sentinel-1 post image and 'Orbit' is ascending(Asc) or descending(Des) orbit information of the Sentinel-1 image. }
\begin{center}
\resizebox{\columnwidth}{!}{%
\begin{tabular}{|c|c|c|c|c|c|}
\hline
\textbf{Tile Ref.}   & \textbf{Site} & \textbf{S1 Post Date} & \textbf{GT Date} & \textbf{Rel. Orbit} & \textbf{Orbit}\\ \hline
\multicolumn{6}{|c|}{Sen1Floods11}\\ \hline
1     & Sri-Lanka & 2017-05-30   & 2017-05-30 & 19         & Des \\ \hline
2     & Slovakia  & 2020-10-20   & 2020-10-20 & 73         & Asc \\ \hline
3     & Somalia   & 2020-05-07   & 2020-05-07 & 116        & Asc  \\ \hline
4     & Spain     & 2019-09-17   & 2019-09-17 & 110        & Des \\ \hline
5,6     & Bolivia  & 2018-02-15  & 2018-02-15 & 156        & Des \\ \hline
7,8     & Mekong  & 2018-08-05   & 2018-08-05 & 26         & Asc  \\ \hline
\multicolumn{6}{|c|}{CEMS} \\ \hline
9,10     & Bosnia    & 2022-04-06   & 2022-04-03 & 51         & Des \\ \hline
11,12,13     & Australia & 2022-03-31   & 2022-03-31 & 147        & Des \\  \hline
14,15     & Scotland & 2022-11-18   & 2022-11-18 & 30        & Asc \\  \hline
\end{tabular}%
}
\label{Flood_data_spec}
\end{center}
\end{table}

\subsection{Data Collection}
\label{subsec:data_collection}

\textbf {Sen1Floods11 Dataset} consist of Sentinel-1 data from 11 different flood events covering a wide variety of geographical area. In total, there are 446 non-overlapped Sentinel-1 tiles in the dataset and each tile is of 512x512 pixel size with a 20-meter ground resolution. Each data sample is composed of two bands VV~(vertical transmit, vertical receive) and VH~(vertical transmit, horizontal receive).
The dataset also provides pixel-wise classification ground truth~(flood segmentation maps). The dataset is hand labeled by experts by using information from Sentinel-1 and Sentinel-2 data followed by manual validation. Each pixel in the ground truth is classified into three categories, 0, 1, and -1. Class 0 represents the absence of water, class 1 represents water, and -1 indicates missing data. Since the ground truth was prepared using both Sentinel-1 and Sentinel-2, heavy clouds in the Sentinel-2 data affected the ground truth preparation. Wherever there is a cloudy pixel in the Sentinel-2, the corresponding pixel in ground truth is marked as missing data i.e., -1. Even though the dataset is big in terms of number of sites covered and number of tiles, a large part of the covered area has no(missing) corresponding ground truth. Therefore, this dataset is not sufficient and we still need a bigger dataset to efficiently train a deep network in a supervised setting. As of now, to the best of our knowledge Sen1Floods11 dataset is the biggest global dataset available on Sentinel-1, hence we decided to use some of good sites for evaluating our unsupervised CD method. To detect, evaluate and visualize flood on each pixel of the region, we choose reliable sites from the test data where there is no missing data in the ground truth. With this criteria, we ended up with tiles from six sites Bolivia, Spain, Cambodia, Slovakia, Somalia, and Sri Lanka.

In Sentinel-1 data samples in the Sen1Floods11 dataset were acquired after floods and are sufficient for a segmentation task. While the CD task requires samples from both pre and post-flood events. Therefore, we also collected pre-flood images and the data collection process is as follows. First, we extracted geometry, relative orbit, and the passing orbit of post-flood images; Second, we downloaded Sentinel-1 images in two months window before the flood event date using geometry and orbit criteria. For each flood event, four pre-flood images~(pre-images) were selected. 
Further data specifications for each site are provided in Table \ref{Flood_data_spec}. All Sentinel-1 data was collected from Google Earth Engine (GEE) \cite{GORELICK201718}. Before downloading, data were pre-processed as described in subsection \ref{subsec:satellite_data_preprocessing}.

\textbf {Copernicus Emergency Management Service~(CEMS)} is one of the six worldwide services provided by the Copernicus program. It provides early warning, monitoring platforms, and mapping services for different natural and man-made disasters. CEMS helps countries with prevention, preparation, response, and recovery activities.
We evaluated our proposed method on three recent flood events listed on the CEMS website under "List of EMS Rapid Mapping Activations"\cite{CEMS_emergency_list}. These floods occurred in current year~(2022) in Mostar, Bosnia \cite{CEMS_flood_Bosnia}, Coraki, Australia \cite{CEMS_flood_Aus} and Aberdeen, Scotland \cite{CEMS_flood_Scot}. The CEMS provides the official flood maps and is publicly available on the CEMS website. According to the information given on the CEMS website, the flood maps for the mentioned flood events were derived from pre and post-event satellite images using a semi-automatic approach. The Bosnia flood map was generated using pre-image from Sentinel-2B and a postimage from the RADARSAT2 satellite. The Australia flood map was generated using pre-image from ESRI imagery and post-image from COSMO-SkyMed satellite. The Scotland flood map was generated using pre-image from ESRI imagery and post-image from Sentinel-1 satellite. The acquisition date of post image which is used in preparing the reference label is mentioned in Table \ref{Flood_data_spec} under column 'GT Date'.

In this study, the reference labels for the three flood event were collected from the CEMS website. These are the official flood maps and publicly accessible. We downloaded and preprocessed Sentinel-1 data for both pre and post-flood events. Multiple tiles were selected for each flood event covering urban, and surrounding agricultural areas. The tile size was kept as 512x512 pixels to maintain consistency with the Sen1Floods11 data. We selected four pre-images for each post-flood tile. Similar to Sen1Floods11 Dataset, all Sentinel-1 data are preprocessed, downloaded from GEE. For further details on collected Sentinel-1 data see Table \ref{Flood_data_spec}.

\begin{figure*}  
\centering  
\begin{subfigure}
  \centering  
  \includegraphics[trim=0cm 0cm 0cm 0cm, width=155mm, height=85mm]{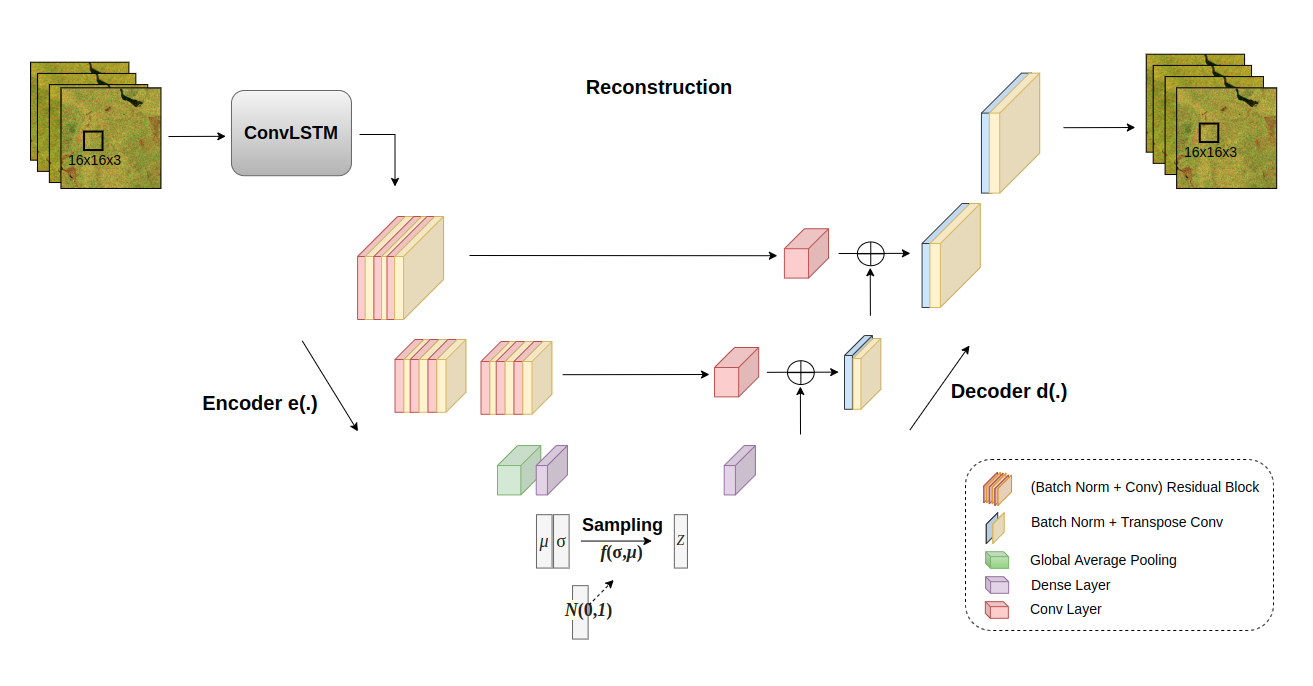}
\end{subfigure}
\caption{Overview of the proposed Network Architecture for unsupervised CD. The network is trained on small 16x16x3 patches to learn the distribution of the small region at a time.}
\label{fig1}
\end{figure*}

\subsection{Data Preprocessing}
\label{subsec:satellite_data_preprocessing}
All collected data were preprocessed and subsequently exported from the cloud-based platform GEE. It is becoming one of the most popular platforms for geospatial big data analysis. One of the biggest advantages of GEE is that Sentinel-1 SAR data is directly available as analysis-ready data cubes. Several studies have highlighted the potential of GEE platform to analyse large amount of geospatial data in a timely manner (e.g.\cite{liu2018high}, \cite{gong2020annual}, \cite{zhang2020development}, \cite{goldblatt2018using}, \cite{ravanelli}). 

The Sentinel-1 mission collects C-band SAR images at 20 m resolution with dual polarization (HH+HV and VV+VH). Sentinel-1 images in GEE were preprocessed to Ground Range Detected (GRD) images using the Sentinel-1 Toolbox. Preprocessing includes removal of thermal noise, radiometric calibration, and terrain correction. In addition, backscatter coefficients were converted to decibels using log scaling ($10\log_{10} x$). We fetched dual-band VV+VH scene acquired in Interferometric Wide swath (IW) mode in a given period, orbit and location. While collecting scenes we also filtered them by Ascending and Descending passes due to the strong influence of incidence angle in the backscatter coefficient. We made sure that orbit pass and relative orbit of all pre and post-flood images are in agreement. Scenes were carefully selected, ensuring better data quality. Then we mask backscatter noise by clipping VV and VH channels in the range (-23, 0) dB and (-28, -5) dB respectively. Finally, both channels are normalized in the range [0, 1].

\section{Methodology}
\label{sec:method}
\subsection{Autoencoder}
In supervised settings, a neural network uses labels to learn features of input data. Labels guide the network to learn specific features depending on the target task, such as classification, segmentation, CD, and others. Input features can also be learned in an unsupervised manner and to do so, autoencoders are one of the widely used network categories. Autoencoders are pixel-wise reconstruction networks, which try to reconstruct their input $x$ from a learned representation $z$. Unlike supervised CNNs, autoencoders generally learn input features for the reconstruction task and can be used for anomaly detection as a downstream task. The architecture of an autoencoder composed of an encoder $e(.)$ [$z= e(x)$], which tries to capture input features $x$ and encodes them into a smaller feature representation $z$; a decoder $d(.)$ [$\hat{x}= d(z)$], which decodes the representation $z$ to reconstruct the input $x$. The network is trained with a loss derived by comparing the input $x$ and the reconstructed output $\hat{x}$.

For our proposed CD network we used a probabilistic autoencoder known as variational autoencoder ($VAE$)\cite{kingma2013auto}. It is a type of generative model that, unlike standard autoencoders, uses probabilistic encoding and decoding and learns to output a distribution over the latent representation $\hat{z}\sim P(z|x)$ and the reconstruction $\hat{x}\sim P(x|z)$. Both encoder and decoder networks in a $VAE$ learn to output the parameters of the learnt distributions, $P(z|x)$ and $P(x|z)$ (e.g., mean and variance in case of a normality assumption). Thus, a trained VAE can be used for both latent representation of an observed input (using the probabilistic encoder) and for generating new unseen data (using the probabilistic decoder). In this work, we propose to detect the changes using the latent representation of a VAE, since it concisely summarizes the content of an input. In particular, we used a divergence measure between the latent distributions (or parameters thereof) of two corresponding image patches to determine whether there is a substantial change from one patch to the other. 

\subsection{Convolutional LSTM}
Long Short-Term Memory (LSTM) is a commonly used method to learn temporal features of a time series data. LSTM however operates on 1xN dimensional vector and does not explore spatial feature learning. For CD tasks on time series data, both spatial and temporal characteristics are significant. Therefore we embedded our proposed network with Convolutional LSTM~(ConvLSTM)~\cite{shi2015convolutional}, an LSTM which captures spatiotemporal correlation. ConvLSTM replaced all 1-d matrix multiplications by convolutional operation thus taking care of spatial neighboring features~\cite{sun2020unet}.


\subsection{Proposed Network Architecture}
\label{sec:Network}
Our training network is shown in Figure \ref{fig1}. The architecture is composed of an encoder, intermediate layers, and a decoder. 
The encoder consists of a convolutional LSTM layer, two residual blocks, a GlobalAveragePooling3D layer, and a dense layer. The convolutional LSTM layer takes a time series of input patches and extracts both temporal and spatial features. Each residual block has three sets of 3D convolutional layer and batch normalization layer. All convolutional layers used kernel of size $3$ and stride $2$. Non-linearity is added using the relu activation function. The residual block is followed by a GlobalAveragePooling3D layer, which calculates the spatial average value for each channel and reduces the dimensionality effectively. At last, a bottleneck dense layer of $8$ channels is added.
The output of the dense layer goes through two intermediate layers which are dense layers of a size equivalent to latent space. We fixed the size of the latent space to $128$. These two dense layers output mean~($\mu_x$) and log-variance~($\sigma_x$) values of the latent distribution corresponding to an input $x$. The output $\mu$ and $\sigma$ of the dense layers are 1D vectors, which are then used to sample a latent vector 
$z \sim  \mathcal{N}(\mu_{x}, \sigma_{x})$
with the help of reparameterization trick \cite{kingma2013auto}, \cite{rezende2014stochastic} for the forward pass to remain differentiable w.r.t. to $\mu$ and $\sigma$.

The decoder takes sampled $z$ as input and passes through a dense layer. These features are then fed into three sets of transpose convolution and batch normalization layers. For transpose convolution Conv3DTranspose layer is implemented. First Conv3DTranspose layer is implemented with kernel size $3$ and stride $2$. Second Conv3DTranspose layer with $3$ filters of kernel size $3x3$, stride $3$, and $'same'$ padding is employed to reconstruct the input stack of patches. In our implementation, we follow the common setup of outputting and using only the mean from the decoder (assuming a fixed unit variance). Furthermore, the decoder network's capacity is improved by employing cross-connections from the encoder network. Note that such skip layers deviate from the standard VAE by having the reconstruction conditioned not only on the latent representation but also on intermediate representations of the encoder. We found this change to be helpful.

The architecture of our proposed method is lightweight since we are using a limited number of time-series images to train the network. Another reason is that Sentinel-1 SAR data is low resolution in comparison to computer vision images and sparse spatial features can be easily learned by a shallow network.

\begin{figure*}[h]
    \centering
    \includegraphics[trim=1cm 0.2cm 1cm 1cm, width=175mm, height=85mm]{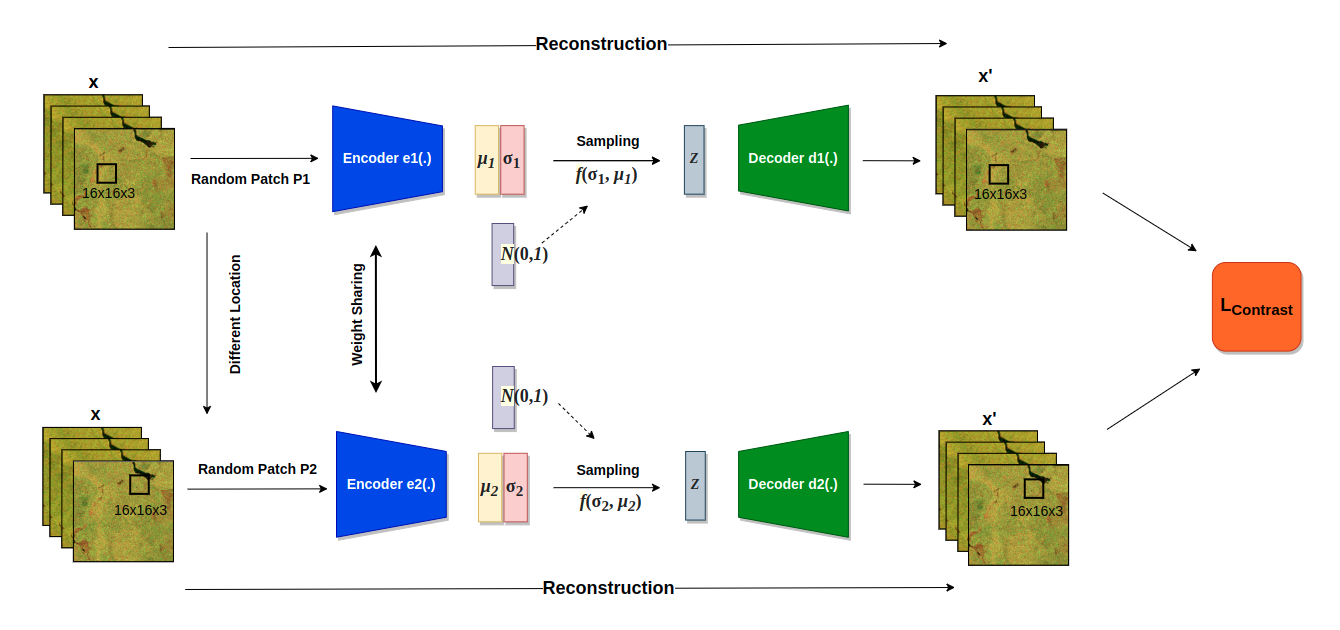}
    \caption{The overview of our unsupervised training pipeline. The training pipeline is to train the model on reconstruction task. Figure Best viewed in color.}
    \label{fig2}
\end{figure*}

\subsection{Training Pipleline}
An overview of the network training pipeline is shown in Figure \ref{fig2}. In the training pipeline, two proposed networks are placed in parallel forming two streams. The inputs to the two streams are time series patches $P1$ and $P2$. The input patches were selected randomly to ensure that they refer to different locations. Both streams were trained to reconstruct their corresponding inputs. Since the two inputs were from different locations, the two networks were also trained to increase the distance between the reconstructed outputs.

\subsection{Training Objective Function}
\label{sec:Objective_function}
The network is optimized using three unsupervised loss functions: a reconstruction loss \cite{kingma2013auto}($L_{Recon}$), a Kullback-Leibler ($KL$) divergence loss \cite{kingma2013auto}($L_{KL}$), and a contrastive loss \cite{hadsell2006dimensionality}($L_{Contrast}$). Note that the first two loss functions constitute the standard VAE objective.


The reconstruction loss encourages the latent representation to contain adequate input information for an accurate reconstruction. The $KL$ divergence loss pushes the latent distribution to be decorrelated and closer to a standard Gaussian. 

The contrastive loss ensures diversity in the reconstructions of the two independent patches $P1$ and $P2$. The patches are from different locations capturing different areas and therefore should most frequently contain dissimilar features. Such a contrastive loss enables the network to learn latent representations that can differentiate features of separate patches. Moreover, it can learn uniformly distributed noise, resulting in a denoising architecture \cite{dong2021residual}. In this work we are proposing an architecture for change detection on SAR data. It is well known that SAR data contains peculiar speckle noise and removing such noise is one of the major challenge in remote sensing. Therefore, adding denoising ability to the architecture is of high importance. The combined objective of the training network is given in the below equation.
\begin{equation}
\label{objective_function}
\begin{aligned}
    L_{Total}   \\
                &= \alpha * \textbf{[} L_{KL} (\mu_{P1}, \sigma_{P1}, N(0,1)) \\
                & + L_{KL} (\mu_{P2}, \sigma_{P2}, N(0,1))\textbf{]}\\
                & + \beta * \textbf{[}L_{Recon} (P1, \hat{P1}) + L_{Recon} (P2, \hat{P2})\textbf{]}\\
                & + (1-\alpha - \beta) * L_{Contrast} (\hat{P1}, \hat{P2})
\end{aligned}
\end{equation}         
where $L_{KL}$ is the $KL$ divergence loss, $L_{Recon}$ is the reconstruction loss. Parameters $\alpha$, $\beta$ are the weight parameter for prioritizing losses. 

\begin{figure*}[h]
    \centering
    \includegraphics[trim=1cm 0.2cm 1cm 1cm, width=165mm, height=72mm]{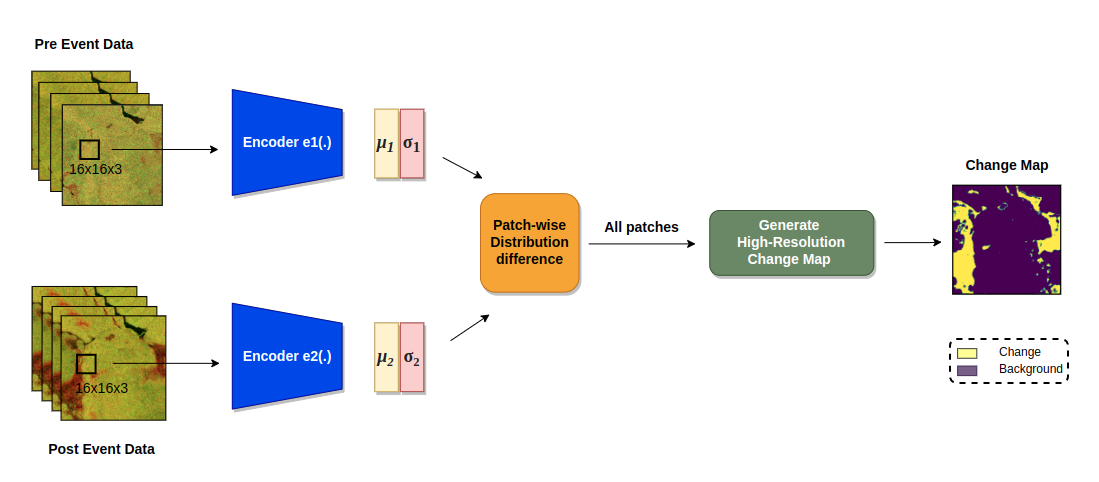}
    \caption{The overview of our inference pipeline. The inference pipeline is for generating change maps between two time stamps. Figure Best viewed in color.}
    \label{fig3}
\end{figure*}

\subsection{Inference}
\label{sec:changeDetect}

Importantly, the proposed network is trained only on pre-event time series data. This means it has adapted its parameters to the distribution of features of pre-event data. Based on this, we assume that the distribution of latent variables for a patch $P1$ should undergo a significant change when affected by an extreme event. This assumption leads us to adopt a simple mechanism to detect change.

The proposed mechanism for CD is shown in Figure \ref{fig3} and algorithm \ref{Algo1}. The inference pipeline used two encoders $e1(.)$ and $e2(.)$ from the trained network. Since the encoders are trained on time series data of length four, we need to provide data of same length while taking the inference. The pre-event input data is prepared by stacking four sequential pre-images(time series), whereas post-event data is formed by stacking one post image four times. This is because Sentinel-1 provides one image every 6 days and flood extension change(increase or decrease) every day. So we use the latest possible image (stacked four times) to get best estimate of the flood extension rather than using four post flood image where the area might not be flooded anymore. Both pre and post-event data is now divided into small patches of size 16x16x3 with stride 1.

From the trained network, the learned distribution can be retrieved as 1D vectors of mean $\mu$ and log-variance for latent representations. The variance$\sigma$ is obtained by taking the exponent of log-variance. We pass pre-event patches through the encoder $e1(.)$ and obtain $\mu _{1}$ , $\sigma _{1}$. Similarly, we pass the post-event patches through encoder $e2(.)$ and retrieve $\mu _{2}$ and $\sigma _{2}$. We can use a number of different measures of divergence or difference between the two distributions of $\mathcal{N}(\mu_1, \sigma_1)$, and $\mathcal{N}(\mu_2, \sigma_2)$ to calculate the change. A few distribution difference measures are tested and compared (see experiment section \ref{dis_diff}). We opt for using a Cosine difference~(CosD) between the $\mu _{1}$ and $\mu _{2}$, defined as follows

\begin{equation}
\label{eq:CD}
    CosD(\mu_{1},\mu_{2})= -\frac{\mu_{1}}{||\mu_{1}||} . \frac{\mu_{2}}{||\mu_{2}||}
\end{equation}

Unlike training, input patches in the two encoders are from the same location. Input patches are generated with stride one and processed with a batch of size 512. The patch-wise distribution difference is calculated using CosD, resulting in a change map ( see COSINE\_DIFF\_MAP in algorithm \ref{Algo1}). Since, the output distribution difference is one value for each 16x16 size patch, the size of change map is smaller in comparison to width and height (W, H) of the input image. We tackled this problem by padding the input pre and post-event images. We used reflect padding or mirroring with length 8 which means reflect padding by 8 pixels on all four sides. This problem could also be resolved by zero padding, but zero padding introduced boundary errors to the detection results \cite{huang2018tiling}. At last, a threshold of -0.9 is applied to get the binary change map (see BIN\_CMAP in algorithm \ref{Algo1}).

\begin{algorithm}
 \KwInput{\textit{Time series pre flood images of length 4(PRE\_IMAGES), post flood image(POST\_IMAGE), PATCH\_SIZE=nx16x16x3, PAD\_SIZE= 8, MODEL}}
 \KwOutput{\textit{Binary change map BIN\_CMAP.}}
  \textit{Pad PRE\_IMAGES and POST\_IMAGE using reflect mode and PAD\_SIZE.}
  
  \textit{PRE\_EVENT\_DATA = stack all PRE\_IMAGES .} 
  
  \textit{POST\_EVENT\_DATA = stack POST\_IMAGE four times.}
  
  \textit{PRE\_PATCH = patches from PRE\_EVENT\_DATA of patchsize and stride 1.}
  
  \textit{POST\_PATCH = patches from POST\_EVENT\_DATA of patchsize and stride 1.}
 
  \For{PRE\_PATCH and POST\_PATCH}
  {
   \textit{PRE\_MEAN, PRE\_STD $=$ MODEL.encoder1.predict(PRE\_PATCH)}
   \textit{POST\_MEAN, POST\_STD $=$ MODEL.encoder2.predict(POST\_PATCH)}
   \textit{Calculate Cosine difference between PRE\_MEAN and POST\_MEAN.}
   }
    \textit{From patch-wise cosine difference get the change map COSINE\_DIFF\_MAP.}
    
    \textit{Apply threshold of -0.9 to get binary change map BIN\_CMAP= COSINE\_DIFF\_MAP$>$-0.9}
    
 \caption{Binary Change Map Inference.}
\label{Algo1}
\end{algorithm}

\begin{figure*}  
\centering  
\begin{subfigure}
  \centering  
  \includegraphics[trim=1cm 0.2cm 1cm 1cm, width=170mm, height=75mm]{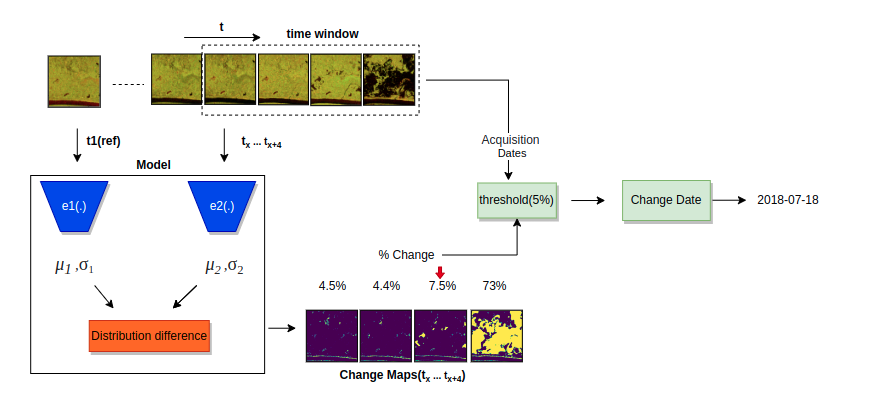}
\end{subfigure}
\caption{Framework for the time-series Change Point Detection. It takes a selected pre-image as reference $x_{ref}$ and generates change maps for all the images in a selected time window and then finds the change point in the time window.}
\label{fig4}
\end{figure*}

\subsection{Change Point Detection}
\label{sec:changePoint}
Change over an area can occur slowly over time such as slow flooding or it can be a sudden event. In case of slow changes, we should identify when the change starts becoming significant. The change can be verified on ground. If valid, the area can then be monitored with priority for further event prediction and warnings. 
With this motive, we developed a change point detection framework where we identify the point(i.e date) when the change is first started appearing on the available SAR data. Change point detection is performed over a long time series and the change is detected compared to a reference image $x_{ref}$. Any pre-image from the training data can be selected as a reference image and the corresponding acquisition date is the reference date $t1_{ref}$. The reference image $x_{ref}$ is the input to the first encoder $e1(.)$.
For the time series data, we select a time window by providing the start date and length of the time window. The length of the time series can be adjusted as per the requirement, for the demonstration we selected the length as 4 and the time series is referred as $t_{x} ... t_{x+4}$. All images from the calculated time window are fetched and processed one by one through encoder $e2(.)$. Change maps are generated for each image from the time window. This is done by following algorithm \ref{Algo1}, where inputs are the $x_{ref}$ and image from the time window (one at a time following the sequence). As our proposed network is currently set to 4 timestamps, both the inputs are stacked four times before feeding into encoder $e1(.)$ and $e2(.)$.

After getting the change maps, a pixel-based percentage change is calculated for each map. The change is considered significant if the percentage change is above a set threshold value. When the threshold is crossed for the first time, the framework fetches the acquisition date of the corresponding input image. The retrieved date is named as change point, i.e. the date when the change started appearing on the SAR data and probably the actual starting point of the change. 
The threshold can be different according to the sensitivity of the targeted change and can be adjusted by changing the parameter. In this study we set the threshold value as median of percentage changes. The overview of the proposed change point detection is depicted in Figure \ref{fig4}.

\subsection{Implementation Details}
One encoder of our proposed network takes four SAR time series images and each SAR here contains three channels. The first 2 channels are VV and VH, whereas the third channel is empty. We are using the three channel network because of two reasons. First, with three channels we can use imagenet pre-trained weights and second, the proposed network architecture can be reused for detecting changes on optical or RGB images.

Our model was trained on a pre-images to learn the distribution of the area and before feeding the images to the training model, input images were split into small patches of size 16x16x3. We employed data augmentation technique to introduce more variations in the data which in turn increase the robustness of the model. This technique is widely used in classification, segmentation, change detection, and other tasks.

We utilized four types of augmentation for our dataset: gaussian blur, gammaContrast, flips and rotation. Enhancing the training dataset with these basic operations improves the performance of CNNs in remote sensing scene classification compared to training on the original dataset \cite{yu2017deep}. Gaussian blur is a non-geometric augmentation that is applied to our input data with a kernel size of 3x3. We use gammaContrast with range (0.25, 2.0) to adjust the image contrast. Both flips and rotation are geometric augmentation methods. Flips were applied left-right with a probability of 0.5 and up-down flips were applied with a probability of 0.2. Rotation was implemented randomly between -90 and 90 degrees. On each input sample, a random combination of Gaussian blur, flips, and rotation was applied before feeding them to the model.

The weight parameter of the objective function $\alpha$ is set to 0.1 which is the weight for $KL$ divergence loss term to bring latent distribution closer to standard normal distribution. $\beta$ is set to 0.7 and given higher weight to reconstruction loss term for accurate reconstruction learning. The remaining 0.2 weight is assigned to contrastive loss term. 
For better convergence, the model was trained with a decaying learning rate. The initial learning rate is 0.001 and decayed until it was at 0.00001. The decay steps were controlled with the "$reduce \ on \ plateau$" method, which decays the rate when the learning curve is stuck at a plateau. The learning rate decay after 2 steps of no learning~(no change in loss) and the training terminates after four steps of no learning. The training network is shallow and lightweight. It contains total 576,395 trainable parameters, therefore the network is faster to train. The network was trained for 10 epochs. All the experiments were implemented in Keras and the training was conducted on one Google Colab GPU. The Code will be released as free and open source and publicly available on our GitHub account soon.

\subsection{Evaluation metric}
The output of the inference network is a pixel-level binary change map. So, the results are evaluated using pixel-level metrics. We used four accuracy metrics namely precision~(P) and recall~(R), F1 score and IoU. The formulas of the metrics are given in Equation \ref{eq:precision} -- \ref{eq:IoU}, where TP represents true positives, FP represents false positives, and FN represents false negatives.

\begin{equation}
\label{eq:precision}
    \text{Precision~(P)} = \frac{\text{TP}}{\text{TP} + \text{FP}}
\end{equation}

\begin{equation}
\label{eq:recall}
    \text{Recall~(R)} = \frac{\text{TP}}{\text{TP} + \text{FN}}
\end{equation}

\begin{equation}
\label{eq:f1_score}
    \text{F1 score} = \frac{\text{TP}}{\text{TP} + \frac{1}{2}(\text{FP} + \text{FN})}
\end{equation}

\begin{equation}
\label{eq:IoU}
    \text{IoU} = \frac{\text{TP}}{\text{TP} + \text{FP} + \text{FN}}
\end{equation}
\\
F1 score metric is the harmonic mean of precision(P) and recall(R) where P and R are given by \ref{eq:precision} and \ref{eq:recall} respectively. The IoU metric measures the intersection over union where TP is the intersection term and union term is (TP + FP + FN). Both F1 score and IoU range from 0 to 1.

\section{Results}
\label{sec:results}

\begin{table*}[h]
  \caption{Quantitative Comparison with Unsupervised Methods: Comparison of our CLAVE method with unsupervised methods log-ratio(Yen), CVA and RaVAEn. The comparison is presented in terms of percentage Recall(R), Precision(P), F1 score(F1), IoU metric. The presented metric values were averaged over 3 runs.}
  \label{quantitative results overview1}
  \centering
  \resizebox{\textwidth}{!}{%
\begin{tabular}{ccccc|cccc|cccc|cccc}

\hline 
\\

\multicolumn{1}{c}{} & \multicolumn{4}{c}{  \textbf{Log-Ratio(Yen's)}} & \multicolumn{4}{c}{ \textbf{CVA}} & \multicolumn{4}{c}{ \textbf{RaVAEn}}& \multicolumn{4}{c}{ \textbf{CLVAE(Ours)}} \\
 
\cline{2-17}
 &  &  &  &  &  &  &  &  &  &  &  &  &  &  &  & \vspace*{-2mm} \\ 
\multicolumn{1}{c}{ \textbf{Site}}   & \textbf{R} & \textbf{P} & \textbf{F1}   & \textbf{IoU} & \textbf{R} & \textbf{P} & \textbf{F1} & \textbf{IoU} & \textbf{R} & \textbf{P} & \textbf{F1} & \textbf{IoU} & \textbf{R} & \textbf{P} & \textbf{F1}    & \textbf{IoU}   \\ \hline
 &  &  &  &  &  &  &  &  &  &  &  &  &  &  &  & \vspace*{-2mm} \\ 
Sri-Lanka (1)  & 31.4       & 11.0       & \textbf{16.3} & \textbf{8.9} & 57.8       & 5.3        & 9.7         & 5.1          & 55.65      & 3.57       & 6.71        & 3.47         & 25.0       & 8.5        & 12.7           & 6.8            \\
Slovakia (2)   & 60.1       & 72.9       & 65.9          & 49.1         & 77.0       & 84.4       & 80.5        & 67.4         & 76.53      & 50.31      & 60.71       & 43.59        & 77.9       & 93.8       & \textbf{85.1}  & \textbf{74.1}  \\
Somalia (3)    & 65.3       & 71.9       & 68.4          & 52.0         & 64.2       & 70.4       & 67.2        & 50.6         & 71.36      & 55.67      & 62.55       & 45.51        & 82.9       & 72.1       & \textbf{77.1}  & \textbf{62.7}  \\
Spain (4)      & 57.4       & 60.4       & 58.9          & 41.7         & 56.4       & 70.4       & 62.6        & 45.6         & 34.97      & 41.91      & 38.13       & 23.55        & 73.8       & 74.7       & \textbf{74.2}  & \textbf{59.0}  \\
Bolivia (5)    & 57.7       & 96.0       & 72.1          & 56.4         & 75.5       & 95.7       & 84.4        & 73.0         & 94.10      & 78.46      & 85.57       & 74.78        & 92.8       & 91.9       & \textbf{92.3}  & \textbf{85.8}  \\
Bolivia (6)    & 34.2       & 92.6       & 50.0          & 33.3         & 63.7       & 60.1       & 61.8        & 44.8         & 65.44      & 78.60      & 71.42       & 55.54        & 81.9       & 81.8       & \textbf{81.8}  & \textbf{69.3}  \\
Mekong (7)     & 59.8       & 93.4       & 72.9          & 57.4         & 76.8       & 99.7       & 86.8        & 76.6         & 67.07      & 91.00      & 77.22       & 62.90        & 89.5       & 96.2       & \textbf{92.7}  & \textbf{86.4}  \\
Mekong (8)     & 41.7       & 94.9       & 57.9          & 40.8         & 78.3       & 99.6       & 87.7        & 78.0         & 72.06      & 86.80      & 78.75       & 64.95        & 94.9       & 95.8       & \textbf{95.3}  & \textbf{91.1}  \\
Bosnia (9)     & 67.7       & 23.6       & 35.0          & 21.3         & 60.6       & 11.0       & 18.6        & 10.2         & 33.00      & 26.53      & 29.41       & 17.24        & 50.0       & 51.3       & \textbf{50.6}  & \textbf{33.9}  \\
Bosnia (10)    & 58.5       & 54.1       & 56.2          & 39.1         & 57.7       & 65.7       & 61.4        & 44.3         & 76.61      & 52.06      & 61.99       & 44.92        & 70.5       & 81.1       & \textbf{75.4}  & \textbf{60.5}  \\
Australia (11) & 53.2       & 69.4       & 60.2          & 43.1         & 75.6       & 90.8       & 82.5        & 70.2         & 86.85      & 76.98      & 81.62       & 68.94        & 93.8       & 91.2       & \textbf{92.5}  & \textbf{86.0}  \\
Australia (12) & 55.7       & 58.5       & 57.1          & 39.9         & 75.1       & 83.2       & 78.9        & 65.2         & 82.75      & 64.44      & 72.46       & 56.81        & 88.9       & 86.2       & \textbf{87.5}  & \textbf{77.8}  \\
Australia (13) & 47.1       & 71.8       & 56.9          & 39.7         & 79.7       & 93.7       & 86.1        & 75.6         & 75.06      & 83.47      & 79.04       & 65.35        & 96.4       & 92.2       & \textbf{94.3}  & \textbf{89.2}  \\
Scotland (14)  & 67.47      & 18.37      & 28.88         & 16.87        & 63.9       & 13.25      & 21.95       & 12.33        & 28.38      & 50.57      & 36.36       & 22.22        & 71.68      & 50.95      & \textbf{59.56} & \textbf{42.41} \\
Scotland (15)  & 71.65      & 24.19      & 36.17         & 22.08        & 60.43      & 9.68       & 16.69       & 9.1          & 18.32      & 56.68      & 27.69       & 16.07        & 65.34      & 55.8       & \textbf{60.19} & \textbf{43.05} \\
 &  &  &  &  &  &  &  &  &  &  &  &  &  &  &  & \vspace*{-2mm} \\ 
 \hline
 &  &  &  &  &  &  &  &  &  &  &  &  &  &  &  & \vspace*{-2mm} \\ 
\multicolumn{1}{c}{ \textbf{Average}}  &55.26      & 60.87      & 52.86         & 37.44        & 68.18      & 63.53      & 60.46       & 48.54        & 62.54      & 59.80      & 57.98       & 44.39        & 77.01      & 74.90      & \textbf{75.43} & \textbf{64.53}\\ 
\hline
  \end{tabular}%
  }\end{table*}

\begin{table*}[h]
  \caption{Quantitative Comparison with Supervised Methods: Comparison of our CLAVE method with supervised change detection methods ADS-Net and DAUSAR. The comparison is presented in terms of percentage Recall(R), Precision(P), F1 score(F1), IoU metric. The presented metric values were averaged over 3 runs.}
  \label{quantitative results overview2}
  \centering
  \resizebox{0.78\textwidth}{!}{%
\begin{tabular}{ccccc|cccc|cccc}

\hline 
\\

\multicolumn{1}{c}{} & \multicolumn{4}{c}{\textbf{ADS-Net}}& \multicolumn{4}{c}{\textbf{DAUSAR}} & \multicolumn{4}{c}{\textbf{CLVAE}(Ours)} \\

\cline{2-13}
 &  &  &  &  &  &  &  &  &  &  &  & \vspace*{-2mm} \\ 
\multicolumn{1}{c}{\textbf{Site}}  & \textbf{R} & \textbf{P} & \textbf{F1}    & \textbf{IoU}  & \textbf{R} & \textbf{P} & \textbf{F1}    & \textbf{IoU}  & \textbf{R} & \textbf{P} & \textbf{F1}    & \textbf{IoU}   \\ \hline
 &  &  &  &  &  &  &  &  &  &  &  & \vspace*{-2mm} \\ 
Sri-Lanka (1)  & 13.7  & 30.9  & 18.98          & 10.5          & 43.8  & 12.7  & \textbf{19.69} & \textbf{10.9} & 25.0  & 8.5   & 12.7           & 6.8            \\
Slovakia (2)   & 94.3  & 80.2  & \textbf{86.68} & \textbf{76.5} & 97.7  & 69.7  & 81.4           & 68.6          & 77.9  & 93.8  & 85.1           & 74.1           \\
Somalia (3)    & 64.4  & 84.6  & 73.13          & 57.6          & 99    & 58.6  & 73.6           & 58.3          & 82.9  & 72.1  & \textbf{77.1}  & \textbf{62.7}  \\
Spain (4)      & 50.5  & 90.2  & 64.75          & 47.8          & 90.4  & 55.7  & 68.9           & 52.6          & 73.8  & 74.7  & \textbf{74.2}  & \textbf{59.0}  \\
Bolivia (5)    & 73.7  & 84.8  & 78.86          & 65.1          & 92.8  & 78.5  & 85.1           & 73.9          & 92.8  & 91.9  & \textbf{92.3}  & \textbf{85.8}  \\
Bolivia (6)    & 47.1  & 82.7  & 60.02          & 42.9          & 63.5  & 77.2  & 69.7           & 53.5          & 81.9  & 81.8  & \textbf{81.8}  & \textbf{69.3}  \\
Mekong (7)     & 93.3  & 95.9  & 94.58          & 89.7          & 96.2  & 93.4  & \textbf{94.8}  & \textbf{90}   & 89.5  & 96.2  & 92.7           & 86.4           \\
Mekong (8)     & 92.9  & 97.3  & 95.05          & 90.6          & 97.1  & 95.1  & \textbf{96.1}  & \textbf{92.5} & 94.9  & 95.8  & 95.3           & 91.1           \\
Bosnia (9)     & 23.2  & 75.8  & 35.53          & 21.6          & 98.2  & 15.8  & 27.2           & 15.7          & 50.0  & 51.3  & \textbf{50.6}  & \textbf{33.9}  \\
Bosnia (10)    & 59.2  & 81.4  & 68.55          & 52.1          & 89.6  & 46.9  & 61.6           & 44.4          & 70.5  & 81.1  & \textbf{75.4}  & \textbf{60.5}  \\
Australia (11) & 99.7  & 76.3  & 86.4           & 76.1          & 96.8  & 84.5  & 90.23          & 82.2          & 93.8  & 91.2  & \textbf{92.5}  & \textbf{86.0}  \\
Australia (12) & 99    & 61.5  & 75.9           & 61.2          & 93.2  & 75.6  & 83.48          & 71.6          & 88.9  & 86.2  & \textbf{87.5}  & \textbf{77.8}  \\
Australia (13) & 99.8  & 79.6  & 88.6           & 79.5          & 97.9  & 87.4  & 92.35          & 85.8          & 96.4  & 92.2  & \textbf{94.3}  & \textbf{89.2}  \\
Scotland (14)  & 38.7  & 45.4  & 41.78          & 26.4          & 81.9  & 35.1  & 49.14          & 32.6          & 71.68 & 50.95 & \textbf{59.56} & \textbf{42.41} \\
Scotland (15)  & 15.9  & 68.2  & 25.79          & 14.8          & 61.8  & 31.7  & 41.9           & 26.5          & 65.34 & 55.8  & \textbf{60.19} & \textbf{43.05} \\
 &  &  &  &  &  &  &  &  &  &  &  &  \vspace*{-2mm} \\ 
 \hline
 &  &  &  &  &  &  &  &  &  &  &  & \vspace*{-2mm} \\ 
\textbf{Avergae}        & 64.36      & 75.65      & 66.31          & 54.16         & 86.66      & 61.19      & 69.01          & 57.27         & 77.01      & 74.90      & \textbf{75.43} & \textbf{64.53}\\ \hline
  \end{tabular}%
  }\end{table*}

\subsection{Compared Methods}
To demonstrate the benefits of our proposed unsupervised CD method~(CLVAE), we compared it with log-ratio and Change Vector Analysis~(CVA) \cite{malila1980change} which are two well-known methods for CD on SAR images. We also show comparison with a recent VAE based unsupervised CD method RaVAEn \cite{ruuvzivcka2021unsupervised}. It is important to note that our method is unsupervised and should be compared with only unsupervised methods. But a good part (6 out of 9 sites) of our tested sites comes from the Sen1Floods11 dataset test set. Therefore, we choose to compare our results with the results produced by the benchmark method on the Sen1Floods11 dataset i.e., the published work with highest IoU score on Sen1Floods11 dataset. To the best of our knowledge DAUSAR \cite{yadav2022attentive} network has provided the highest score on the Sen1Floods11 dataset, therefore in our work we will refer to this work as the benchmark method. We also compared with a recent deeply supervised network ADS-Net. We want to emphasize that the motive of the comparison is not only to find the best performing method, but also to study generalizability of our proposed novel unsupervised CD network. Below is an overview of the compared methods. 
\begin{enumerate}

\item Log-ratio is commonly used to highlight changes in pairs of bi-temporal SAR images (e.g., \cite{hu2014unsupervised}), and is formally defined as follows: 

\begin{equation}
\begin{small}
\label{eq:log_ratio}
    \text{LR} = 10 \log_{10} \left(\frac{S_{tn}}{S_{t1}}\right) = 10\log_{10}(S_{tn}) - 10\log_{10}(S_{t1})
\end{small}
\end{equation}
where $S_{t1}$ and $S_{tn}$ are the SAR images acquired over the same geographical area at the beginning and end of the time series, respectively.
However, prior to computing the log-ratio, we used the Lee filter to suppress speckle noise from both images \cite{lee1981speckle}. Binary change maps from log-ratio were generated using both otsu\cite{otsu1979threshold} and Yen's thresholding methods \cite{yen1995new}.
\item CVA is a popular CD method for multispectral optical images and SAR images. CVA generates change magnitude and change direction separately, which can be useful in determining change areas and change types. In this work, we focus on the magnitude of the change. Therefore, we calculated the magnitude change using CVA and then the changes are binarized using otsu thresholding.

\item RaVAEn is an unsupervised method recently proposed to detect changes in Sentinel-2 multispectral images instead of SAR. For comparison, we adapted RaVAEn on SAR data. This method uses a simple VAE network with residual encoder trained using default VAE losses. The method uses 32x32 size patches and change on each patch is calculated using cosine difference. Generated changes maps contain pixelated changes (see Figures \ref{fig:SOTA_QUAL} and \ref{fig:SOTA_QUAL2}) that result in low-resolution change maps.

\item DAUSAR is a supervised benchmark network \cite{yadav2022attentive} on Sen1Floods11 SAR dataset. It is a deep convolutional network for pixel-wise detailed CD between pre and post event images. Since the original Sen1Floods11 contains only post flood images, the authors extended the dataset by adding pre flood Sentinel-1 images. The pre-flood images were collected from GEE. The network is dual-stream Siamese U-Net enhanced with spatial and channel-wise attention. 

\item ADS-Net is a supervised change detection network. It is a deep convolutional network which uses multi-scale features to extract changes between bitemporal remote sensing images. ADS-Net proved better change detection compared to existing deeply supervised networks including the famous FC-Siamese networks among others. This Network is proposed for 3 channel RGB optical images. We implemented the network for SAR images, where we used VV, VH, and VV/VH as three channels. We trained the network on pre-flood and post-flood images.
\end{enumerate}

All mentioned methods are implemented fro scratch. The two supervised network DAUSAR and ADS-Net are trained on the training set of Sen1Floods11 dataset and corresponding Sentinel-1 pre flood images from GEE. We use these trained network to take inference on our test sites listed in Table ~\ref{Flood_data_spec}. The inference results will further be used to see how well a supervised method trained on Sen1Floods11 dataset can generalize to new flood sites from CEMS website.

\subsection{Quantitative Results}
The quantitative comparison of the above-mentioned methods with our CLVAE is presented in Table \ref{quantitative results overview1} and \ref{quantitative results overview2}. The first table shows the comparison with all unsupervised methods and the second table with supervised methods. The comparison is displayed on each site in terms of four accuracy metrics; Recall, Precision, F1 score, and IoU. The log-ratio method with otsu thresholding is performing significantly low in comparison to the log-ratio with Yen's thresholding. Hence in Table \ref{quantitative results overview1} we presented the results from the best performing threshold method. However, the comparison of log-ratio with otsu thresholding is depicted later in Figures \ref{fig:SOTA_QUAL} and \ref{fig:SOTA_QUAL2} for qualitative analysis. 

\begin{figure}[htbp]
\centerline{\includegraphics[width=92mm, height=68mm]{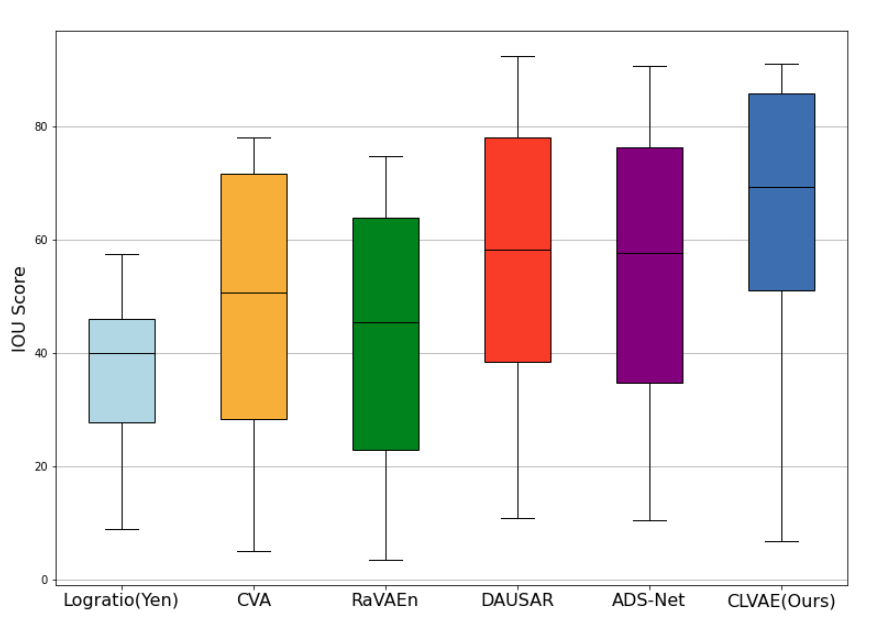}}
\caption{Boxplot Graph: Graphical IoU comparison of CLVAE with log-ratio with Yen's thresholding, CVA, RaVAEn, DAUSAR and ADS-Net change detection methods.}
\label{SOTA_BoXPLOT}
\end{figure}

\begin{figure}[htbp]
\centerline{\includegraphics[width=90mm, height=68mm]{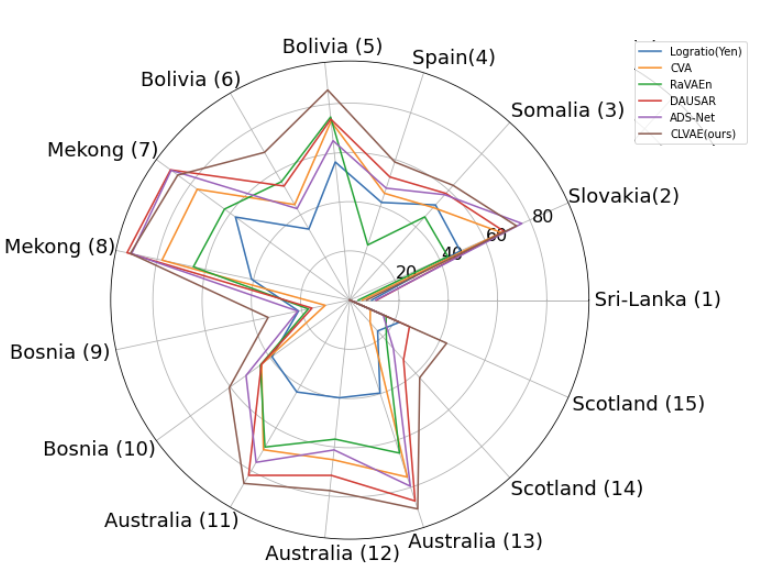}}
\caption{Spider Graph: Graphical IoU comparison of our method with log-ratio with Yen's thresholding, CVA, RaVAEn, DAUSAR and ADS-Net change detection methods.}
\label{SOTA_SPIDER}
\end{figure}

On average, CLVAE achieved an F1 score of 75.43\% and IoU score of 64.53\% with 77.01\% recall and 74.90\% precision. Among all the compared methods ours achieved the best average precision, F1 score and IoU, whereas recall is best achieved by the DAUSAR. In comparison to the unsupervised CD methods log-ratio, CVA and RaVAEn, CLVAE outperformed in all four average metrics. The lead in recall ranges from 9-22\%, in precision from 11-15\%, in F1 score from 15-22\% and in IoU from 16-27\%. RaVAEn was originally proposed for CD on multi spectral data and seems to be not promising for SAR data. This method performed better then log-ratio but couldn't outperform other compared methods. The supervised method DAUSAR has a high recall percentage(86.66\%) which is approximately 10\% higher compared to CLAVE's recall. Also, ADS-Net 75.65\% precision which is 0.7\% better then CLAVE's precision. But overall CLVAE maintains high precision and recall and outperformed the supervised methods by 6\% in F1 and 7\% in IoU score. 

On individual sites, CLVAE yielded the best F1 score and IoU except for 'Slovakia', 'Mekong' and 'Sri-Lanka' sites. On 'Slovakia' site DAS-Net gave the best results and it's DAUSAR on 'Mekong' site. Compared to CLVAE scores, the difference is not much and ranges from 1-2\% in F1 score and 1-4\% in IoU metric. On 'Sri-Lanka' site, DAUSAR gave the best results but the scores are extremely low. The site is a rice field that was flooded right after harvesting months. Therefore the half-cut stems are good enough to hold flood water and change is reflected between pre, post images. However, these flooded fields were not considered flooded in the ground truth leading to disagreement between the detected change and the ground truth. This explains the low scores by CLVAE and all compared methods.

On most of the sites recall is best achieved by the supervised method DAUSAR. Since high recall of DAUSAR is associated with a low precision, the IoU and F1-scores are also relatively low. Another minor deviation from the average metric results is in precision on 'Bolivia' and 'Mekong' sites. Unlike average results, the precision on 'Bolivia' sites is best by log-ratio and on 'Mekong' sites by CVA. But they also have low recall which leads them to significantly low IoU and F1 scores.

Further insights into the results are provided by Figure \ref{SOTA_BoXPLOT} and \ref{SOTA_SPIDER} where the IoU results are visualized in boxplot and spider graphs. In the boxplot, the x-axis represents the compared method and the y-axis represents the percentage IoU score. Notably, our CLVAE gave the highest IoU score median. In the spider graph, the axis represents the evaluation sites and the numbers on all the concentric circles represent the possible percentage IoU score from 0 at the center to 100 on the outermost circle. The farther toward the end of the spike, the larger the value. Closest to the center means closer to zero. The outermost line represents the best performing model and in the current scenario it is our proposed method CLVAE. 

\begin{figure*}
     
     \begin{subfigure}
         \centering
         \includegraphics[trim=0cm 0cm 0cm 0cm, width=180mm, height=75mm]{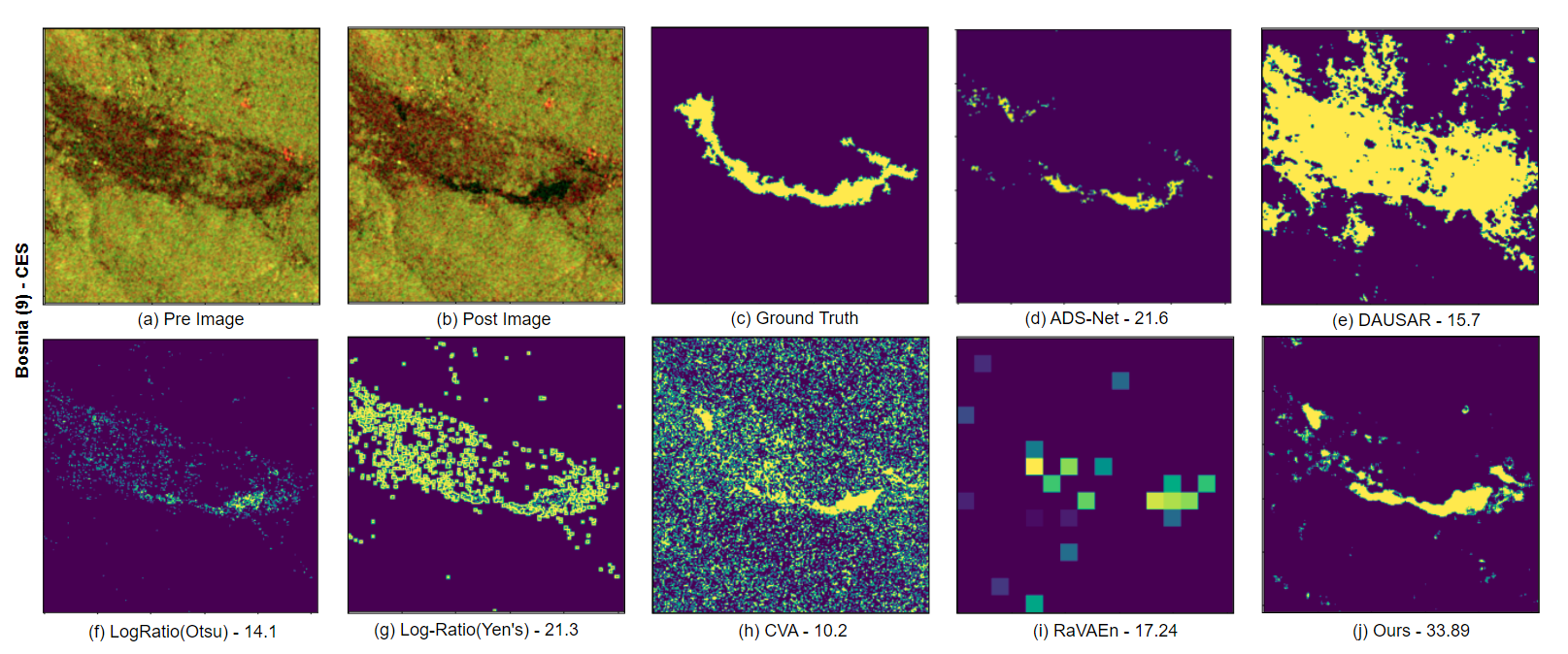}
     \end{subfigure}
     \begin{subfigure}
         \centering
         \includegraphics[trim=0cm 0cm 0cm 0cm, width=180mm, height=75mm]{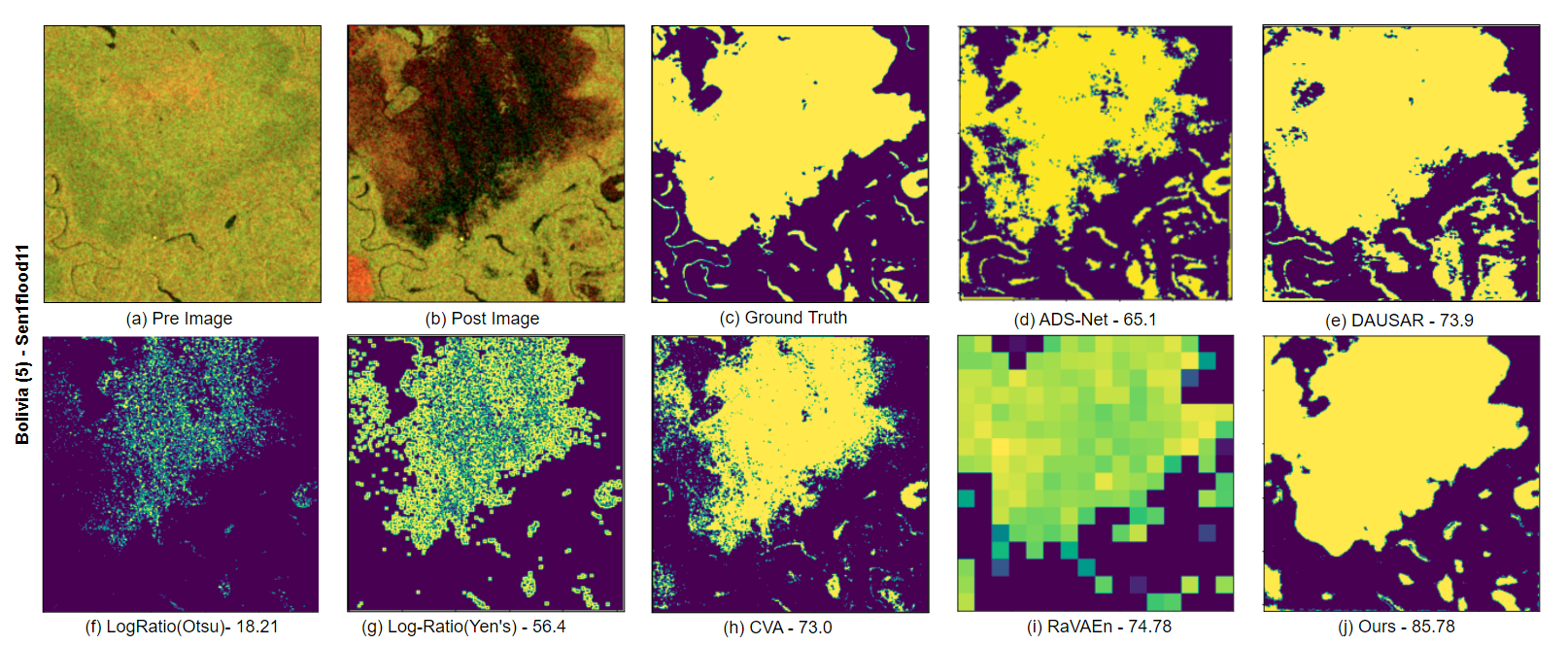}
     \end{subfigure}
     \centering
     \begin{subfigure}
         \centering
         \includegraphics[trim=0cm 0cm 0cm 0cm, width=180mm, height=75mm]{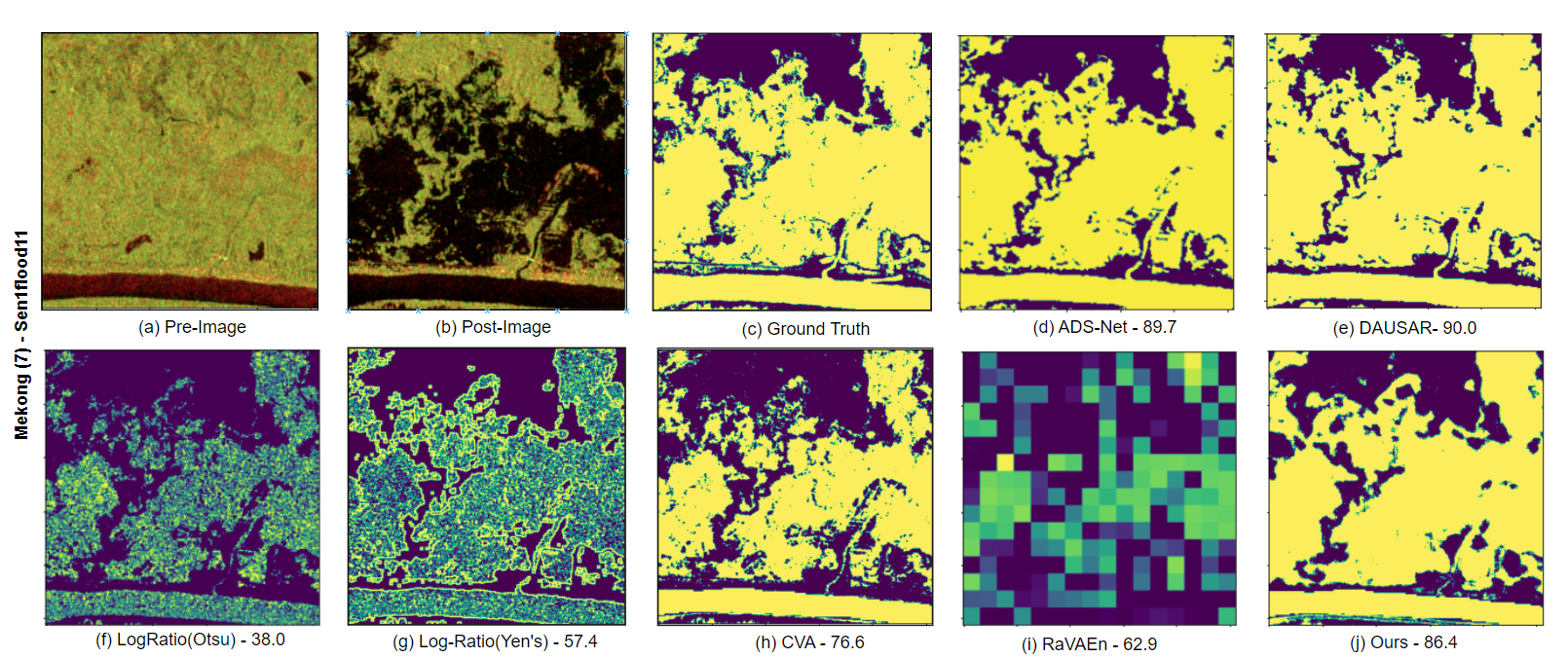}
     \end{subfigure}
\caption{Qualitative Comparison: (a), (b) and (c) represents latest pre flood image, post flood image and binary ground truth. The remaining images shows comparison of (d) ADS-Net, (e) DAUSAR, (f) Log-ratio with otsu threshold, (g) Log-ratio with yen's threshold, (h) CVA, and (i) RaVAEn with (j) our proposed CLVAE CD method. The number below each change map is corresponding percentage IoU score.}
\label{fig:SOTA_QUAL}%
\end{figure*}

\begin{figure*}

     \centering
     \begin{subfigure}
         \centering
         \includegraphics[trim=0cm 0cm 0cm 0cm, width=185mm, height=75mm]{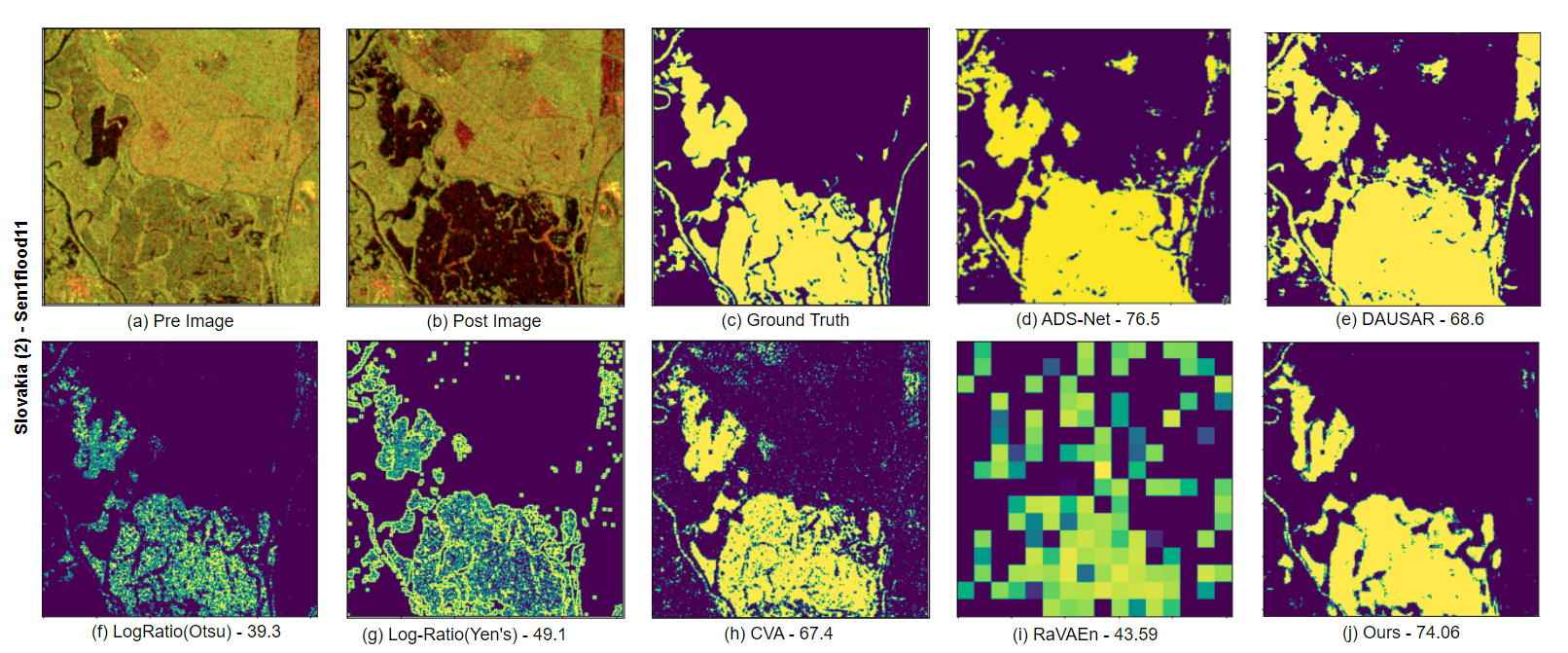}
     \end{subfigure}
     \begin{subfigure}
         \centering
         \includegraphics[trim=0cm 0cm 0cm 0cm, width=185mm, height=75mm]{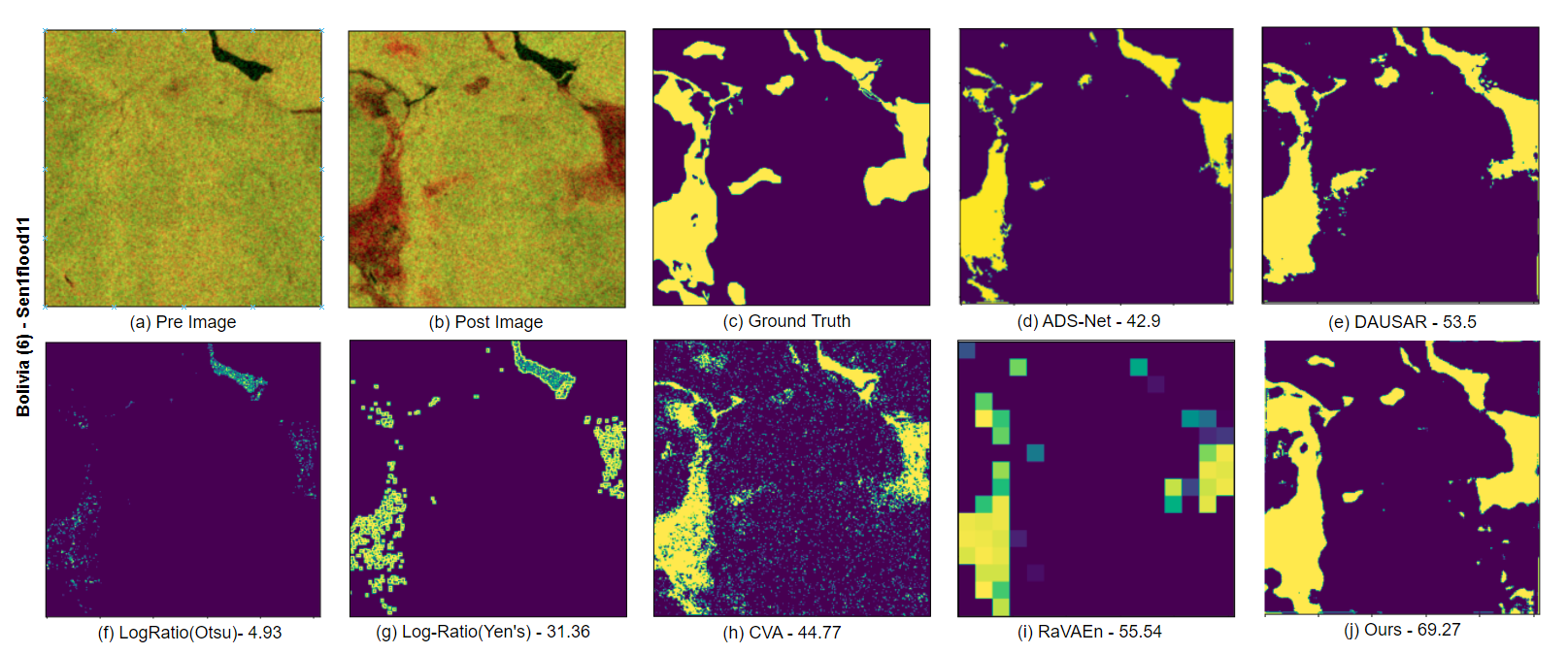}
     \end{subfigure}
\caption{Qualitative Comparison: (a), (b) and (c) represents latest pre flood image, post flood image and binary ground truth. The remaining images shows comparison of (d) ADS-Net, (e) DAUSAR, (f) Log-ratio with otsu threshold, (g) Log-ratio with yen's threshold, (h) CVA, and (i) RaVAEn with (j) our proposed CLVAE CD method. The number below each change map is corresponding percentage IoU score.}
\label{fig:SOTA_QUAL2}%
\end{figure*}

\subsection{Qualitative Results}
For qualitative analysis, we selected three geographically different and challenging sites; Mekong, Bolivia, Slovakia and Bosnia. The visualization of change maps is presented in Figure \ref{fig:SOTA_QUAL} and \ref{fig:SOTA_QUAL2}, where (a) and (b) show the latest pre-flood image and post-flood image, (c) show the ground truth, (d) and (e) show the change map from supervised methods ADS-Net and DAUSAR, (f), (g), (h), (i) and (j) show the change map generated from log-ratio with otsu threshold, log-ratio with Yen's threshold, CVA, RaVAEn and our proposed CLVAE respectively. The number below each change map is the corresponding percentage IoU score.

The First two rows of Figure \ref{fig:SOTA_QUAL} show the detection results on 'Bosnia' site. Log-ratio detected a good part of the flooded area correctly. But the detection is not smooth rather grainy causing false negatives. The speckle noise in surrounding areas is also adding false positives. CVA detected a good portion of the flooded area but the change map contains high speckle noise. RaVAEn detected changes in 32x32 patches showing ill-defined pixelated change. The change map doesn't contain major false detection but also failed to detect most part of the changes shown in ground truth. Our CLVAE also couldn't detect the flooded area with huge success but provided significantly good detection in comparison to others. The generated change map contains very low false detection and did not suffer from speckle noise. It is noteworthy that the supervised method ADS-Net is missing a good part of the flooded area and DUASAR suffers from a large amount of false detection whereas our unsupervised method CLVAE produced relatively good detection results. In terms of IoU, CLVAE gave 11 to 23\% better score than the compared methods.

The third and fourth rows of Figure \ref{fig:SOTA_QUAL} shows the detection results on 'Bolivia' site. The detection on this site is comparatively better than the 'Bosnia' site. CVA shows good detection results and doesn't suffer from speckle noise. However, both log-ratio and CVA missed a significant part of the flooded area(change). The detection from RaVAEn is also better but lacks details due to the coarse resolution of the output change maps. Both supervised methods show good detection results but ADS-Net suffers from false negatives and DAUSAR suffers from false positives. Compared to the six methods our CLVAE method resulted in a clear and speckle-free change map. On 'Bolivia' site, the IoU score of CLVAE is 10 to 30\% better than the compared methods.

The last two rows of Figure \ref{fig:SOTA_QUAL} show the detection results on 'Mekong' site. All the CD methods performed well on this site. The supervised methods ADS-Net and DAUSAR outperformed unsupervised CD methods. Among unsupervised results, log-ratio has problem of grainy detections, both CVA and RaVAEn missed some of the changed areas, RaVAEn suffers from pixelated detection and CLVAE missed small flooded streams. On an average all methods detected majority of the flooded areas.

Two more samples from 'Slovakia' and 'Bolivia' sites are are shown for evaluation in Figure \ref{fig:SOTA_QUAL2}. The first two rows shows the detection results on 'Slovakia' site, where ADS-Net gave better(2\% better IoU score) detection results compared to CLVAE. Whereas, the second sample from 'Bolivia' site shows that best results are provided by our CLVAE.



\section{Experiments}
\label{sec:experiments}

\subsection{Performance With Respect to different Distribution Difference Methods}
\label{dis_diff}
As discussed in Section~\ref{sec:changeDetect}, with our proposed CD architecture,  different choices of difference between $\mathcal{N}_1=\mathcal{N}(\mu_{1},\sigma_{1})$ and $\mathcal{N}_2=\mathcal{N}(\mu_{2},\sigma_{2})$ are possible for change detection. Here, we tested four different methods for calculating distribution difference, namely Kullback-Leibler Divergence~(KLD), Jensen-Shannon Divergence~(JSD), Euclidean Distance~(ED) and Cosine Distance~(CosD). The KL and JSD methods operate on full distribution (using both mean and variance), whereas we use ED and CosD only on the mean parameter of the distribution. The formulas for all four distance functions is given in Eq. \ref{eq:KLD} -- \ref{eq:CD2}, where $P$, $Q$ are the distributions, $\mu$ represents mean, $\sigma$ is variance and $\left\| \right\|$ represents L2 normalization function.

\begin{alignat}{5}
    &KLD(\mathcal{N}_1\lVert\mathcal{N}_2) =\quad\;\;\;&& \sum_{}^{}log(\frac{\sigma_{2}}{\sigma_{1}}) + \frac{\sigma_{1}^{2} + (\mu_{1}-\mu_{2})^{2}}{2\sigma_{2}^{2}} \label{eq:KLD} \\
    &JSD(\mathcal{N}_1\lVert\mathcal{N}_2) 
        =&& \frac{KLD(\mathcal{N}_1\lVert\mathcal{N}_m)}{2} 
        +\frac{KLD(\mathcal{N}_1\lVert\mathcal{N}_m)}{2} \label{eq:JSD} \\
    &\qquad\qquad\small\text{where, }&&\mathcal{N}_m = \mathcal{N}(\frac{\mu_{1} +\mu_{2}}{2}, \frac{(\sigma_{1} +\sigma_{2})}{2}) \nonumber \\
    &ED(\mathcal{N}_1,\mathcal{N}_2)=&&\sqrt{\sum_{}^{}(\mu_{1}-\mu_{2})^{2}} \label{eq:ED} \\
    &CosD(\mathcal{N}_1,\mathcal{N}_2)=&& -\frac{\mu_{1}}{\left\| \mu_{1} \right\|}\, . \,\frac{\mu_{2}}{\left\| \mu_{2} \right\|} \label{eq:CD2}
\end{alignat}

The average metric calculated by the mentioned four methods is shown in Table \ref{tab:distribution_diff_method} where CosD shows the best mean precision, F1 score and IoU value. The recall with CosD function is lower in comparison to other compared functions. 
It is important to point out that KLD, JSD, and ED resulted in similar values. This can be explained as follows, as explained in subsections \ref{sec:Network} and \ref{sec:Objective_function}, in the training process, our network is guided to learn input data as standard normal distribution encouraging the variance to be 1. As a result, the majority of variance values from the trained encoder, at inference time, turn out to be 1 as well. This is true for both pre and post-flood images. If we equate $\sigma_{1}$ and $\sigma_{2}$ to 1 in KLD eq. \ref{eq:KLD}, it comes down to $\frac{1}{2}\sum_{}^{}(\mu_{1}-\mu_{2})^{2}$. At this point, both ED and KL values are some positive fraction of $\sum_{}^{}(\mu_{1}-\mu_{2})^{2}$. In the calculation of distribution difference \ref{Algo1}, the threshold for KLD, JSD, and ED is set to 0.0 indicating that only the existence of change is considered and not the magnitude of change. This condition equalizes the change maps from KLD and ED methods. The same explanation is valid for results with the JSD method as well.
Therefore, we end up with ED and CosD as two different difference measures. On average, both distribution difference functions show similar results but these values are significantly different on individual sites. 

\begin{table}[h]
\caption{Performance variation with respect to Distribution difference Functions.}
\label{tab:distribution_diff_method}
\centering
\resizebox{.40\textwidth}{!}{%
\begin{tabular}{ccccc}
\hline
\multicolumn{1}{l}{} & Mean R & Mean P & Mean F1 & Mean IoU \\
 \cline{1-5} 

KLD &  \textbf{80.37}	& 69.84	& 73.93	& 62.64 \\
JSD &  \textbf{80.37}	& 69.84	& 73.93	& 62.64 \\
ED & \textbf{80.37}	& 69.84	& 73.93	& 62.64 \\
CosD & 77.01 &	\textbf{74.90}	 & \textbf{75.43} & \textbf{64.53}
\end{tabular}%
  }
\end{table}

\subsection{Performance Variation With Respect to Number of Residual Blocks}
In the encoder part of our proposed network, the first two residual blocks downsample the input data. The third block is for feature learning without downsampling. Before reaching our network architecture settings, we experimented with the number of non-downsampling residual blocks. The average results of the conducted experiments are shown in Table \ref{tab:n_residual blocks}. Best results were recorded with one non-downsampling residual block, therefore this setting is used in our proposed network. Increasing the number of residual blocks further shows a slight decrease in the performance. Also, note that our approach is unsupervised and use limited training data. Therefore a shallow network is an appropriate choice hence the behavior.

\begin{table}[h]
\caption{Performance variation with respect to number of residual blocks.}
\label{tab:n_residual blocks}
\centering
\resizebox{.40\textwidth}{!}{%
\begin{tabular}{ccccc}
\hline
Blocks & Mean R & Mean P & Mean F1 & Mean IoU \\
 \cline{1-5} 
0 & 75.30 & 72.46 & 72.67 & 61.13 \\
1 & \textbf{77.01} &	\textbf{74.90}	 & \textbf{75.43} & \textbf{64.53} \\
2 & 74.97	& 71.75	& 72.17	& 60.26
\end{tabular}%
  }
\end{table}

\begin{figure}[htbp]
\centerline{\includegraphics[width=85mm, height=60mm]{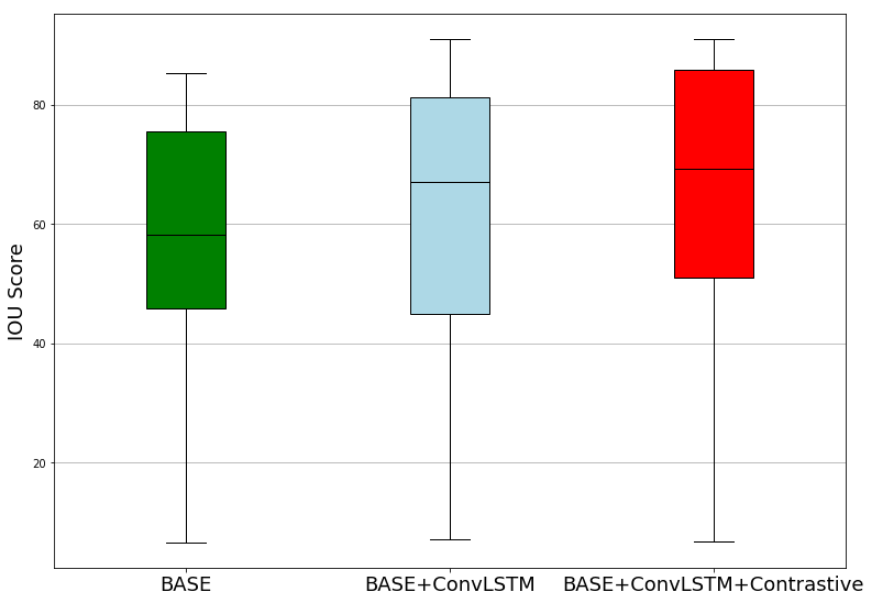}}
\caption{Ablation results visualized with boxplot depicting mean IoU score across all sites. X-axis represents mean IoU score and y-axis represents boxplots for BASE network, BASE with Convolution LSTM network and our proposed CLAVE method which is BASE network with convolutional LSTM and trained with contrastive learning.}
\label{ablation}
\end{figure}

\begin{figure*}

     \centering
     \begin{subfigure}
         \centering
         \includegraphics[trim=0cm 0cm 0cm 0cm, width=136mm, height=93mm]{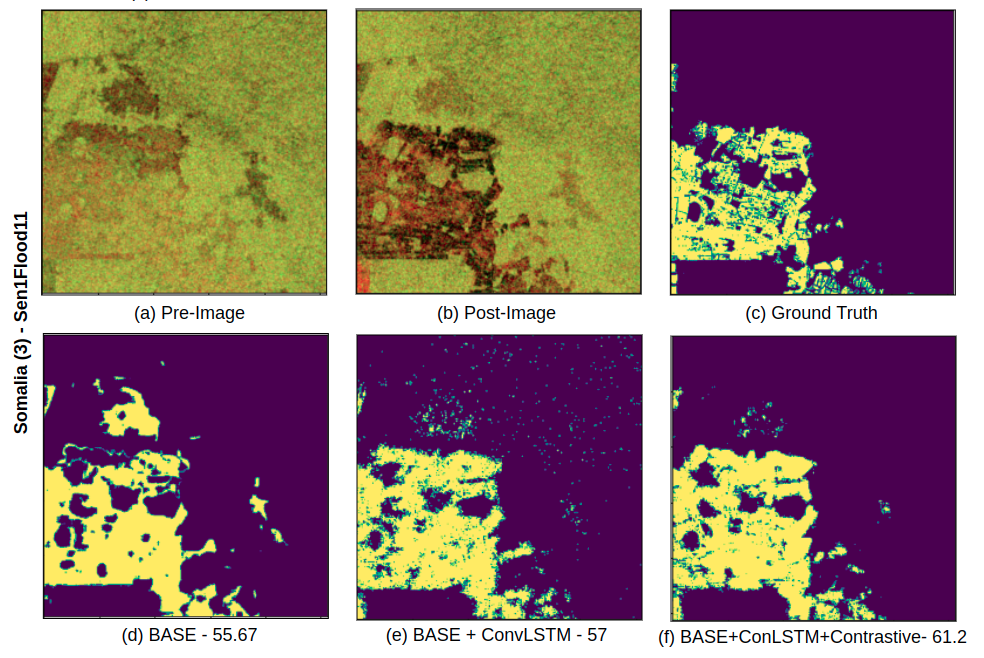}
     \end{subfigure}
     \begin{subfigure}
         \centering
         \includegraphics[trim=0cm 0cm 0cm 0cm, width=135mm, height=93mm]{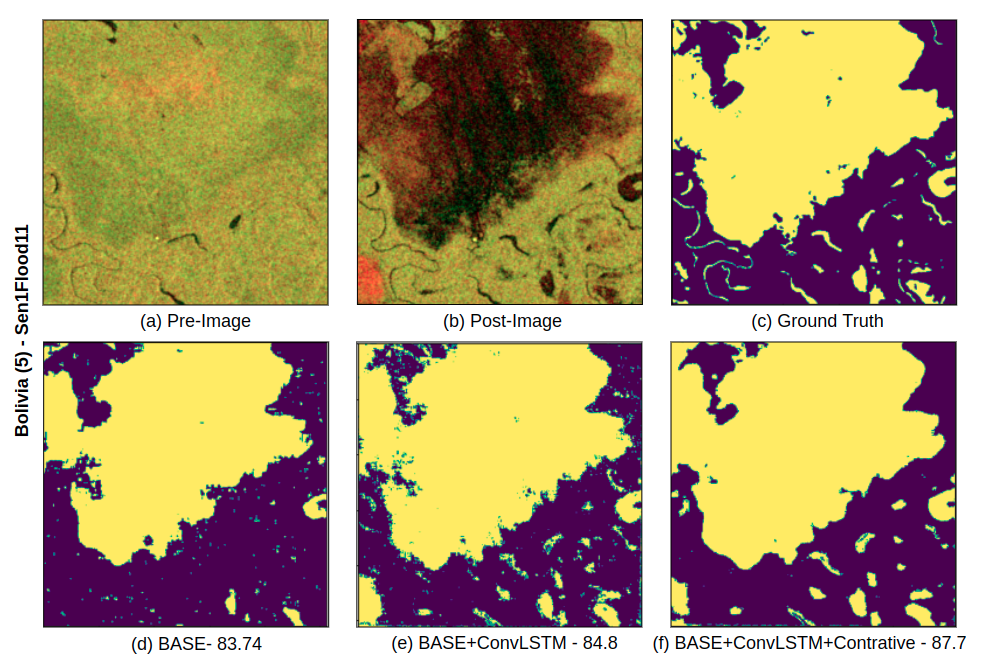}
     \end{subfigure}
\caption{Qualitative Comparison for Ablation Study: In first row (a), (b) and (c) represents latest pre-flood image, post-flood image and ground truth. In second row (d), (e) and (f) represents detection results from BASE network(encoder-decoder based VAE), BASE with Convolutional LSTM, and proposed CLVAE CD method(BASE with Convolutional LSTM trained using contrastive learning). The number below each change map is corresponding percentage IoU score.}
\label{fig:ablation_QUAL}%
\end{figure*}

\subsection{Ablation Experiments}
The term “ablation study” is borrowed from the medical field that consist of removing parts of the nervous system of vertebrates to understand their purpose. This technique was originally introduced by the French physiologist M.J.P. Flourens ~\cite{MJ_Flourens}. In DL, ablation is removal of parts of the network and analysing the performance of the resulting networks. It helps in investigating the contribution of different parts or techniques used in the DL network. In this study we conducted an ablation of convLSTM and contrastive learning method. Quantitative results are shown in Table \ref{tab:ablation_table}. 

\begin{table}[h]
\caption{Ablation Study. Comparison of BASE architecture with Convolutional LSTM, BASE architecture with contrastive learning and our proposed CLVAE network.}
\label{tab:ablation_table}
\centering
\resizebox{.5\textwidth}{!}{%
\begin{tabular}{ccccc}
\hline
\multicolumn{1}{l}{} & Mean R & Mean P & Mean F1 & Mean IoU \\
 \cline{1-5} 

BASE & 74.75 &	69.99 &	70.87 &	58.44 \\
+ ConvLSTM & 76.90 &	73.95 &	73.68 &	61.89 \\
+ Contrastive Learning & \textbf{77.01} &	\textbf{74.90}	 & \textbf{75.43} & \textbf{64.53} 			
\end{tabular}%
  }
\end{table}




The 'BASE' network refers to the encoder-decoder based VAE reconstruction network. The four metric values given in the table are averaged over all the sites. The results depict that convLSTM helped the network to learn better representation and lead to better recall, precision, F1 score, and IoU. Contrastive learning, on the other hand, improved the precision of the results at the risk of lower recall. Contrastive learning also provided better F1 and IoU score, indicating that training the model with contrastive learning reduce false detections~(FN and FP). Figure \ref{ablation} gives further insight into the ablation study in form of box plots depicting IoU score on y-axis. From left to right we see an increase in median value or increase in length of the upper quartile, which indicates an increase in IoU score for at least 50\% of the sites.

For the qualitative analysis of the effect of ConvLSTM and contrastive learning can be seen in the two samples shown in Figure \ref{fig:ablation_QUAL}. In the first sample from Bolivia, ConvLSTM (e) detected more changes but also added speckle noise in some regions. The speckle-noise is then removed by contrastive learning (f). In the second sample from 'Somalia' site, ConvLSTM (e) eliminates false detections and gave more true positives at the risk of small speckle noise. The speckle noise is then removed with the help of contrastive learning(see (f)), leading to a better IoU score.

ConvLSTM helps in learning better feature representation and can detect the changes more efficiently. But at the same time, it introduces speckle noise to the result, which is handled by training the model with contrastive learning. With contrastive learning, our model learns speckle noise and avoids that to be part of a change. This in turn helps the framework identify meaningful differences between pre and post-flood images.

\subsection{Performance Variation With Respect to Time Series Length }
In this section, we show how our proposed CD network performed with smaller and bigger time series data. Before selecting the settings of the proposed CLVAE model we experimented with different time series lengths. The average quantitative results corresponding to 2, 4, and 8 time series lengths are given in Table \ref{tab:time_series}. Our network's performance improved as we increased the time series length from two to four. We see a small decline when the network is supplied with eight pre-images. The observed reason is an increase in seasonal changes and frequent partial flooding. Our network gave the best detection results with pre-images from the same season. It is noteworthy that, at the cost of a small decline in performance of the network is still reliable to use for a longer time series.

\begin{table}[h]
\caption{Performance variation with respect to length(2, 4 and 8) of time series.}
\label{tab:time_series}
\centering
\resizebox{.40\textwidth}{!}{%
\begin{tabular}{ccccc}
\hline
 & Mean R & Mean P & Mean F1 & Mean IoU \\
 \cline{1-5}
2 & 72.20 & 68.40 & 70.24 & 54.14 \\
4 & 77.01 &	\textbf{74.90}	 & \textbf{75.43} & \textbf{64.53} \\
8 & \textbf{77.19} &71.68 &	73.65 &	62.40
\end{tabular}%
  }
\end{table}

\subsection{Performance Variation With Respect to Patch Size.}
We also experimented with the network's input patch size. The average metric values of the results are shown in Table \ref{tab:patch_size}. The network performed best with patch size 16x16. Our CD network uses patch-wise distribution differences to generate the final change map. As we increase the patch size, the network fails to capture small changes efficiently through the distribution difference. Therefore smaller patch gave better results shown below.

\begin{table}[h]
\caption{Performance variation with respect to input patch size.}
\label{tab:patch_size}
\centering
\resizebox{.44\textwidth}{!}{%
\begin{tabular}{ccccc}
\hline
 & Mean R & Mean P & Mean F1 & Mean IoU \\
 \cline{1-5}
16x16 & \textbf{77.01} &	\textbf{74.90}	 & \textbf{75.43} & \textbf{64.53} \\
32x32 & 76.08	& 69.45	& 71.56	& 59.22
\end{tabular}%
  }
\end{table}

\begin{figure*}
\centerline{\includegraphics[width=187 mm, height=80mm]{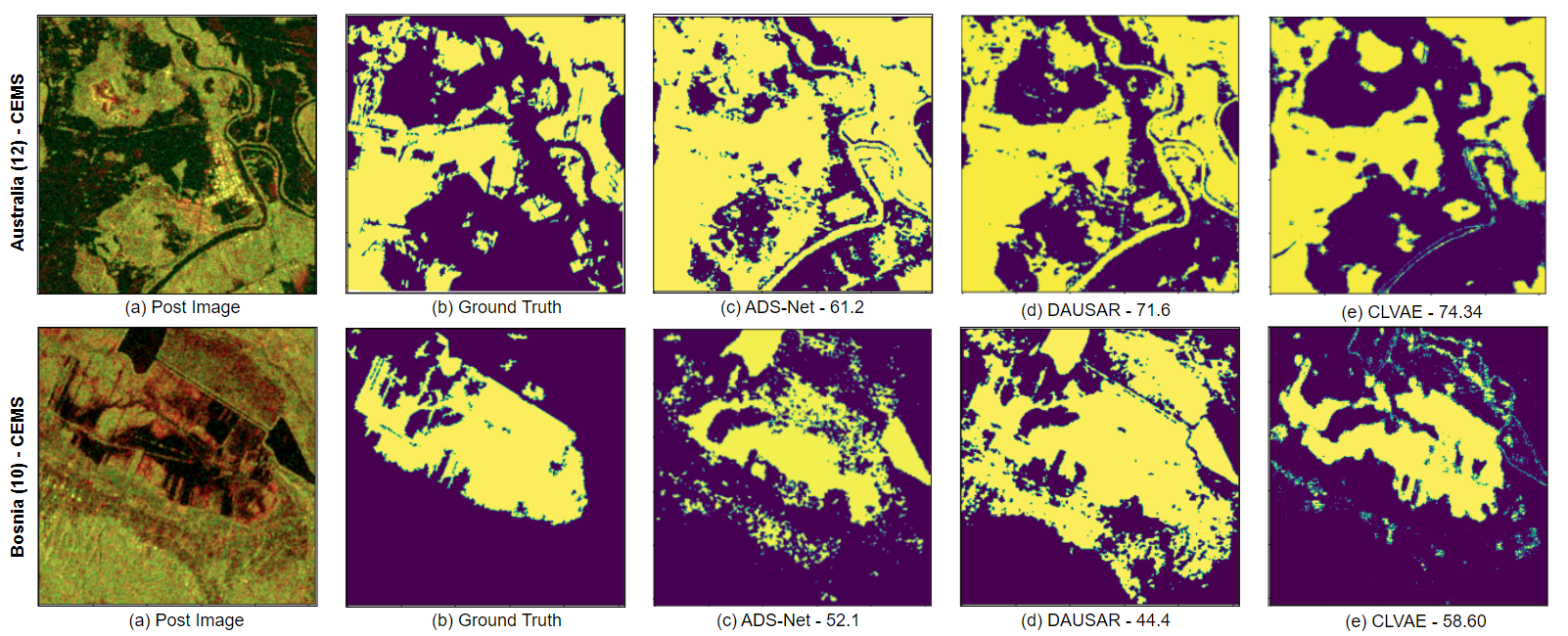}}
\caption{Qualitative Comparison for Generalizability Test: (a) represents post flood image, (b) represents ground truth, (c) represents change map from supervised Sen1Floods11 benchmark method, and (d) represents change map from our proposed CLVAE method. The number below each change map is corresponding percentage IoU score. }
\label{General_samples}
\end{figure*}

\subsection{Generalizability}
The generalizability of a network is its capability to generate good results on unseen sites. Our proposed CLVAE is an unsupervised CD network and hence requires no label for training. This means that we do not need to rely on the generalizability of our model. Rather we can train it on unlabeled SAR data from any area of interest covered by SAR satellite (Sentinel-1) and use it for CD. However, we conducted experiments to investigate the generalizability of our proposed unsupervised CD method compared to the supervised methods.

We trained our CLVAE network only on pre-images from 'Spain' flood site and took inference on all CEMS sites. The generated average results are given in Table \ref{tab:Generalizability}. This experiment shows that our unsupervised method is still performing better than the supervised Sen1Floods11 benchmark method. On average CLVAE gave better precision, F1 score and IoU score. It is also worth noting that, the supervised method DAUSAR gave a high recall but really low precision. Big difference between recall(high) and precision(low) indicates a lot of false positives, which we can see in the qualitative results presented in Figure \ref{General_samples}. ADS-Net on the other hand gave lower recall as well as lower precision compared to our proposed unsupervised change detection method,

\begin{table}[h]
\caption{Generalizability comparison of proposed unsupervised CLVAE method with two supervised methods ADS-Net and DAUSAR trained on Sen1Floods11 benchmark.}
\label{tab:Generalizability}
\centering
\resizebox{.45\textwidth}{!}{%
\begin{tabular}{ccccc}
\hline
\multicolumn{1}{l}{} & Mean R & Mean P & Mean F1 & Mean IoU \\
 \cline{1-5}
CLVAE & 78.49	& \textbf{71.52} & 	\textbf{74.8} & \textbf{59.79}\\
ADS-Net & 62.21	& 69.74 & 	60.36	& 47.39 \\
DAUSAR & \textbf{88.49}	& 53.86	& 63.70	& 51.26 \\

\end{tabular}%
  }
\end{table}

For qualitative comparison, change maps for two flood sites 'Australia' and 'Bosnia' are shown in Figure \ref{General_samples}. In both samples, the supervised method detected a large portion of the non-flooded area as flooded. Therefore generate false positives. Whereas our CLVAE gave significantly better CD results. On 'Australia' site CLVAE outperformed the supervised method by 3 to 13\% and on 'Bosnia' site by 6 to 14\%. Even though we see a drop in CLAVE performance when it is trained on one site and inference is taken on others, the drop is not that high. The CD by CLVAE is still generalizable to other geographically different and unseen sites. 

\section{Conclusion}
\label{sec:conclusion}
In this paper, we proposed a novel unsupervised remote sensing CD method based on a probabilistic model. Our method CLVAE cumulatively benefits from the reconstruction approach, latent parameters learning of probabilistic autoencoder, distribution difference method, convolutional LSTM, and contrastive learning techniques. Our model strongly learns the spatiotemporal correlation between time series SAR data. The extensive experimental results on Sen1Floods11 data and CEMS data display the potential of the proposed CD method. Our method yield 64.53\% average IoU and 75.43\% average F1 score. Our results have surpassed the performance of existing unsupervised non-DL methods i.e. log-ratio, CVA, and unsupervised DL method RaVAEn. On average our CLVAE has a lead of 10-21\% in terms of F1 score and 7-27 \% in terms of IoU score. As 6 out of 9 sites in our evaluation data are from Sen1Floods11 test data, we also compared our results with the supervised methods ADS-Net and DAUSAR trained on Sen1Floods11 dataset. Our unsupervised CD method CLVAE shows a lead of 6\% F1 score and 7\% IoU over compared supervised method.
Further to this, we also presented a change point detection framework based on our CD method. The framework can detect changes at early stages which in turn can save lives through timely evacuation, alerts, and other disaster management activities. In light of new satellite missions, a better temporal frequency of one or two images per day can be immensely helpful in monitoring the change point. The proposed method and framework are light on memory, have low computation time(faster training and inference), and are also inexpensive in terms of data preparation as no annotation is required.

In this study, we proposed a CD method on Sentinel-1 SAR data and presented its efficiency in detecting floods. In future we will test and extend our change detection method in other applications areas as well. Different remote sensing sensors capture specific features of the scene. In our future works we would like to train our network with complimenting features from multi-sensors.
Another potential direction is urban flood detection using high-resolution data, where our model can be trained to detect flooded buildings and roads at a better resolution. This can help local transport agencies to reroute the traffic saving lives of pedestrians and drivers.

\bibliographystyle{IEEEtran}
\bibliography{refer}

\begin{thebibliography}{10}
\providecommand{\url}[1]{#1}
\csname url@samestyle\endcsname
\providecommand{\newblock}{\relax}
\providecommand{\bibinfo}[2]{#2}
\providecommand{\BIBentrySTDinterwordspacing}{\spaceskip=0pt\relax}
\providecommand{\BIBentryALTinterwordstretchfactor}{4}
\providecommand{\BIBentryALTinterwordspacing}{\spaceskip=\fontdimen2\font plus
\BIBentryALTinterwordstretchfactor\fontdimen3\font minus
  \fontdimen4\font\relax}
\providecommand{\BIBforeignlanguage}[2]{{%
\expandafter\ifx\csname l@#1\endcsname\relax
\typeout{** WARNING: IEEEtran.bst: No hyphenation pattern has been}%
\typeout{** loaded for the language `#1'. Using the pattern for}%
\typeout{** the default language instead.}%
\else
\language=\csname l@#1\endcsname
\fi
#2}}
\providecommand{\BIBdecl}{\relax}
\BIBdecl

\bibitem{disaster_report}
``2021 disasters in numbers,''
  \url{https://reliefweb.int/report/world/2021-disasters-numbers#:~:text=In%202021%2C%20a%20total%20of,across%20the%202001%2D2020%20period
  }, accessed: 2022-06-30.

\bibitem{anusha2020flood}
N.~Anusha and B.~Bharathi, ``Flood detection and flood mapping using
  multi-temporal synthetic aperture radar and optical data,'' \emph{The
  Egyptian Journal of Remote Sensing and Space Science}, vol.~23, no.~2, pp.
  207--219, 2020.

\bibitem{di2011timely}
G.~Di~Baldassarre, G.~Schumann, L.~Brandimarte, and P.~Bates, ``Timely low
  resolution sar imagery to support floodplain modelling: a case study
  review,'' \emph{Surveys in geophysics}, vol.~32, no.~3, pp. 255--269, 2011.

\bibitem{lu2004change}
D.~Lu, P.~Mausel, E.~Brondizio, and E.~Moran, ``Change detection techniques,''
  \emph{International journal of remote sensing}, vol.~25, no.~12, pp.
  2365--2401, 2004.

\bibitem{luppino2019unsupervised}
L.~T. Luppino, F.~M. Bianchi, G.~Moser, and S.~N. Anfinsen, ``Unsupervised
  image regression for heterogeneous change detection,'' \emph{arXiv preprint
  arXiv:1909.05948}, 2019.

\bibitem{malila1980change}
W.~A. Malila, ``Change vector analysis: An approach for detecting forest
  changes with landsat,'' in \emph{LARS symposia}, 1980, p. 385.

\bibitem{bovolo2008multilevel}
F.~Bovolo, ``A multilevel parcel-based approach to change detection in very
  high resolution multitemporal images,'' \emph{IEEE Geoscience and Remote
  Sensing Letters}, vol.~6, no.~1, pp. 33--37, 2008.

\bibitem{thonfeld2016robust}
F.~Thonfeld, H.~Feilhauer, M.~Braun, and G.~Menz, ``Robust change vector
  analysis (rcva) for multi-sensor very high resolution optical satellite
  data,'' \emph{International Journal of Applied Earth Observation and
  Geoinformation}, vol.~50, pp. 131--140, 2016.

\bibitem{deng2008pca}
J.~Deng, K.~Wang, Y.~Deng, and G.~Qi, ``Pca-based land-use change detection and
  analysis using multitemporal and multisensor satellite data,''
  \emph{International Journal of Remote Sensing}, vol.~29, no.~16, pp.
  4823--4838, 2008.

\bibitem{dharani2021land}
M.~Dharani and G.~Sreenivasulu, ``Land use and land cover change detection by
  using principal component analysis and morphological operations in remote
  sensing applications,'' \emph{International Journal of Computers and
  Applications}, vol.~43, no.~5, pp. 462--471, 2021.

\bibitem{asokan2019change}
A.~Asokan and J.~Anitha, ``Change detection techniques for remote sensing
  applications: a survey,'' \emph{Earth Science Informatics}, vol.~12, no.~2,
  pp. 143--160, 2019.

\bibitem{daudt2018fully}
R.~C. Daudt, B.~Le~Saux, and A.~Boulch, ``Fully convolutional siamese networks
  for change detection,'' in \emph{2018 25th IEEE International Conference on
  Image Processing (ICIP)}.\hskip 1em plus 0.5em minus 0.4em\relax IEEE, 2018,
  pp. 4063--4067.

\bibitem{bandara2022revisiting}
W.~G.~C. Bandara and V.~M. Patel, ``Revisiting consistency regularization for
  semi-supervised change detection in remote sensing images,'' \emph{arXiv
  preprint arXiv:2204.08454}, 2022.

\bibitem{zhang2020deeply}
C.~Zhang, P.~Yue, D.~Tapete, L.~Jiang, B.~Shangguan, L.~Huang, and G.~Liu, ``A
  deeply supervised image fusion network for change detection in high
  resolution bi-temporal remote sensing images,'' \emph{ISPRS Journal of
  Photogrammetry and Remote Sensing}, vol. 166, pp. 183--200, 2020.

\bibitem{chen2020dasnet}
J.~Chen, Z.~Yuan, J.~Peng, L.~Chen, H.~Huang, J.~Zhu, Y.~Liu, and H.~Li,
  ``Dasnet: Dual attentive fully convolutional siamese networks for change
  detection in high-resolution satellite images,'' \emph{IEEE Journal of
  Selected Topics in Applied Earth Observations and Remote Sensing}, vol.~14,
  pp. 1194--1206, 2020.

\bibitem{wang2021ads}
D.~Wang, X.~Chen, M.~Jiang, S.~Du, B.~Xu, and J.~Wang, ``Ads-net: An
  attention-based deeply supervised network for remote sensing image change
  detection,'' \emph{International Journal of Applied Earth Observation and
  Geoinformation}, vol. 101, p. 102348, 2021.

\bibitem{yadav2022attentive}
R.~Yadav, A.~Nascetti, and Y.~Ban, ``Attentive dual stream siamese u-net for
  flood detection on multi-temporal sentinel-1 data,'' \emph{arXiv preprint
  arXiv:2204.09387}, 2022.

\bibitem{chen2021remote}
H.~Chen, Z.~Qi, and Z.~Shi, ``Remote sensing image change detection with
  transformers,'' \emph{IEEE Transactions on Geoscience and Remote Sensing},
  vol.~60, pp. 1--14, 2021.

\bibitem{munoz2021local}
D.~F. Mu{\~n}oz, P.~Mu{\~n}oz, H.~Moftakhari, and H.~Moradkhani, ``From local
  to regional compound flood mapping with deep learning and data fusion
  techniques,'' \emph{Science of the Total Environment}, vol. 782, p. 146927,
  2021.

\bibitem{manjusree2012optimization}
P.~Manjusree, L.~Prasanna~Kumar, C.~M. Bhatt, G.~S. Rao, and V.~Bhanumurthy,
  ``Optimization of threshold ranges for rapid flood inundation mapping by
  evaluating backscatter profiles of high incidence angle sar images,''
  \emph{International Journal of Disaster Risk Science}, vol.~3, no.~2, pp.
  113--122, 2012.

\bibitem{zhong2017spectral}
Z.~Zhong, J.~Li, Z.~Luo, and M.~Chapman, ``Spectral--spatial residual network
  for hyperspectral image classification: A 3-d deep learning framework,''
  \emph{IEEE Transactions on Geoscience and Remote Sensing}, vol.~56, no.~2,
  pp. 847--858, 2017.

\bibitem{noh2022unsupervised}
H.~Noh, J.~Ju, M.~Seo, J.~Park, and D.-G. Choi, ``Unsupervised change detection
  based on image reconstruction loss,'' in \emph{Proceedings of the IEEE/CVF
  Conference on Computer Vision and Pattern Recognition}, 2022, pp. 1352--1361.

\bibitem{jing2022remote}
W.~Jing, S.~Zhu, P.~Kang, J.~Wang, S.~Cui, G.~Chen, and H.~Song, ``Remote
  sensing change detection based on unsupervised multi-attention slow feature
  analysis,'' \emph{Remote Sensing}, vol.~14, no.~12, p. 2834, 2022.

\bibitem{ruuvzivcka2021unsupervised}
V.~R{\v{z}}i{\v{c}}ka, A.~Vaughan, D.~De~Martini, J.~Fulton, V.~Salvatelli,
  C.~Bridges, G.~Mateo-Garcia, and V.~Zantedeschi, ``Unsupervised change
  detection of extreme events using ml on-board,'' \emph{arXiv preprint
  arXiv:2111.02995}, 2021.

\bibitem{zhan2018log}
T.~Zhan, M.~Gong, X.~Jiang, and S.~Li, ``Log-based transformation feature
  learning for change detection in heterogeneous images,'' \emph{IEEE
  Geoscience and Remote Sensing Letters}, vol.~15, no.~9, pp. 1352--1356, 2018.

\bibitem{liu2016deep}
J.~Liu, M.~Gong, K.~Qin, and P.~Zhang, ``A deep convolutional coupling network
  for change detection based on heterogeneous optical and radar images,''
  \emph{IEEE transactions on neural networks and learning systems}, vol.~29,
  no.~3, pp. 545--559, 2016.

\bibitem{chen2020simple}
T.~Chen, S.~Kornblith, M.~Norouzi, and G.~Hinton, ``A simple framework for
  contrastive learning of visual representations,'' in \emph{International
  conference on machine learning}.\hskip 1em plus 0.5em minus 0.4em\relax PMLR,
  2020, pp. 1597--1607.

\bibitem{chen2020improved}
X.~Chen, H.~Fan, R.~Girshick, and K.~He, ``Improved baselines with momentum
  contrastive learning,'' \emph{arXiv preprint arXiv:2003.04297}, 2020.

\bibitem{grill2020bootstrap}
J.-B. Grill, F.~Strub, F.~Altch{\'e}, C.~Tallec, P.~Richemond, E.~Buchatskaya,
  C.~Doersch, B.~Avila~Pires, Z.~Guo, M.~Gheshlaghi~Azar \emph{et~al.},
  ``Bootstrap your own latent-a new approach to self-supervised learning,''
  \emph{Advances in neural information processing systems}, vol.~33, pp.
  21\,271--21\,284, 2020.

\bibitem{caron2018deep}
M.~Caron, P.~Bojanowski, A.~Joulin, and M.~Douze, ``Deep clustering for
  unsupervised learning of visual features,'' in \emph{Proceedings of the
  European conference on computer vision (ECCV)}, 2018, pp. 132--149.

\bibitem{dong2021multiscale}
H.~Dong, W.~Ma, L.~Jiao, F.~Liu, and L.~Li, ``A multiscale self-attention deep
  clustering for change detection in sar images,'' \emph{IEEE Transactions on
  Geoscience and Remote Sensing}, vol.~60, pp. 1--16, 2021.

\bibitem{saha2021self}
S.~Saha, P.~Ebel, and X.~X. Zhu, ``Self-supervised multisensor change
  detection,'' \emph{IEEE Transactions on Geoscience and Remote Sensing},
  vol.~60, pp. 1--10, 2021.

\bibitem{luppino2022code}
L.~T. Luppino, M.~A. Hansen, M.~Kampffmeyer, F.~M. Bianchi, G.~Moser,
  R.~Jenssen, and S.~N. Anfinsen, ``Code-aligned autoencoders for unsupervised
  change detection in multimodal remote sensing images,'' \emph{IEEE
  Transactions on Neural Networks and Learning Systems}, 2022.

\bibitem{deldari2021time}
S.~Deldari, D.~V. Smith, H.~Xue, and F.~D. Salim, ``Time series change point
  detection with self-supervised contrastive predictive coding,'' in
  \emph{Proceedings of the Web Conference 2021}, 2021, pp. 3124--3135.

\bibitem{bonafilia2020sen1floods11}
D.~Bonafilia, B.~Tellman, T.~Anderson, and E.~Issenberg, ``Sen1floods11: a
  georeferenced dataset to train and test deep learning flood algorithms for
  sentinel-1,'' in \emph{Proceedings of the IEEE/CVF Conference on Computer
  Vision and Pattern Recognition Workshops}, 2020, pp. 210--211.

\bibitem{CEMS}
``Copernicus emergency management services,''
  \url{https://emergency.copernicus.eu/mapping/copernicus-emergency-management-service#zoom=2&lat=31.47858&lon=7.20923&layers=0BT00},
  accessed: 2020-09-30.

\bibitem{GORELICK201718}
\BIBentryALTinterwordspacing
N.~Gorelick, M.~Hancher, M.~Dixon, S.~Ilyushchenko, D.~Thau, and R.~Moore,
  ``Google earth engine: Planetary-scale geospatial analysis for everyone,''
  \emph{Remote Sensing of Environment}, vol. 202, pp. 18--27, 2017, big
  Remotely Sensed Data: tools, applications and experiences. [Online].
  Available:
  \url{https://www.sciencedirect.com/science/article/pii/S0034425717302900}
\BIBentrySTDinterwordspacing

\bibitem{CEMS_emergency_list}
``Copernicus list of ems risk and recovery mapping activations,''
  \url{https://emergency.copernicus.eu/mapping/list-of-activations-risk-and-recovery},
  accessed: 2020-09-30.

\bibitem{CEMS_flood_Bosnia}
``Cems bosnia flood,''
  \url{https://emergency.copernicus.eu/mapping/system/files/components/EMSR572_AOI01_DEL_PRODUCT_r1_RTP02_v1.pdf},
  accessed: 2022-09-30.

\bibitem{CEMS_flood_Aus}
``Cems australia flood,''
  \url{https://emergency.copernicus.eu/mapping/system/files/components/EMSR570_AOI01_DEL_MONIT01_r1_RTP01_v2.pdf},
  accessed: 2022-09-30.

\bibitem{CEMS_flood_Scot}
``Cems scotland flood,''
  \url{https://emergency.copernicus.eu/mapping/system/files/components/EMSR640_AOI02_DEL_MONIT01_r1_RTP02_v3.pdf},
  accessed: 2022-09-30.

\bibitem{liu2018high}
X.~Liu, G.~Hu, Y.~Chen, X.~Li, X.~Xu, S.~Li, F.~Pei, and S.~Wang,
  ``High-resolution multi-temporal mapping of global urban land using {Landsat}
  images based on the {Google} {Earth} {Engine} platform,'' \emph{Remote
  sensing of environment}, vol. 209, pp. 227--239, 2018.

\bibitem{gong2020annual}
P.~Gong, X.~Li, J.~Wang, Y.~Bai, B.~Chen, T.~Hu, X.~Liu, B.~Xu, J.~Yang,
  W.~Zhang \emph{et~al.}, ``Annual maps of global artificial impervious area
  ({GAIA}) between 1985 and 2018,'' \emph{Remote Sensing of Environment}, vol.
  236, p. 111510, 2020.

\bibitem{zhang2020development}
X.~Zhang, L.~Liu, C.~Wu, X.~Chen, Y.~Gao, S.~Xie, and B.~Zhang, ``Development
  of a global 30 m impervious surface map using multisource and multitemporal
  remote sensing datasets with the {Google} {Earth} {Engine} platform,''
  \emph{Earth System Science Data}, vol.~12, no.~3, pp. 1625--1648, 2020.

\bibitem{goldblatt2018using}
R.~Goldblatt, M.~F. Stuhlmacher, B.~Tellman, N.~Clinton, G.~Hanson,
  M.~Georgescu, C.~Wang, F.~Serrano-Candela, A.~K. Khandelwal, W.-H. Cheng
  \emph{et~al.}, ``Using {Landsat} and nighttime lights for supervised
  pixel-based image classification of urban land cover,'' \emph{Remote Sensing
  of Environment}, vol. 205, pp. 253--275, 2018.

\bibitem{ravanelli}
R.~Ravanelli, A.~Nascetti, R.~V. Cirigliano, C.~Di~Rico, G.~Leuzzi, P.~Monti,
  and M.~Crespi, ``Monitoring the impact of land cover change on surface urban
  heat island through {Google} {Earth} {Engine}: Proposal of a global
  methodology, first applications and problems,'' \emph{Remote Sensing},
  vol.~10, no.~9, 2018.

\bibitem{kingma2013auto}
D.~P. Kingma and M.~Welling, ``Auto-encoding variational bayes. corr
  abs/1312.6114 (2013),'' \emph{arXiv preprint arXiv:1312.6114}, vol. 482,
  2013.

\bibitem{shi2015convolutional}
X.~Shi, Z.~Chen, H.~Wang, D.-Y. Yeung, W.-K. Wong, and W.-c. Woo,
  ``Convolutional lstm network: A machine learning approach for precipitation
  nowcasting,'' \emph{Advances in neural information processing systems},
  vol.~28, 2015.

\bibitem{sun2020unet}
S.~Sun, L.~Mu, L.~Wang, and P.~Liu, ``L-unet: An lstm network for remote
  sensing image change detection,'' \emph{IEEE Geoscience and Remote Sensing
  Letters}, 2020.

\bibitem{rezende2014stochastic}
D.~J. Rezende, S.~Mohamed, and D.~Wierstra, ``Stochastic backpropagation and
  approximate inference in deep generative models,'' in \emph{International
  conference on machine learning}.\hskip 1em plus 0.5em minus 0.4em\relax PMLR,
  2014, pp. 1278--1286.

\bibitem{hadsell2006dimensionality}
R.~Hadsell, S.~Chopra, and Y.~LeCun, ``Dimensionality reduction by learning an
  invariant mapping,'' in \emph{2006 IEEE Computer Society Conference on
  Computer Vision and Pattern Recognition (CVPR'06)}, vol.~2.\hskip 1em plus
  0.5em minus 0.4em\relax IEEE, 2006, pp. 1735--1742.

\bibitem{dong2021residual}
N.~Dong, M.~Maggioni, Y.~Yang, E.~P{\'e}rez-Pellitero, A.~Leonardis, and
  S.~McDonagh, ``Residual contrastive learning for joint demosaicking and
  denoising,'' \emph{arXiv preprint arXiv:2106.10070}, 2021.

\bibitem{huang2018tiling}
B.~Huang, D.~Reichman, L.~M. Collins, K.~Bradbury, and J.~M. Malof, ``Tiling
  and stitching segmentation output for remote sensing: Basic challenges and
  recommendations,'' \emph{arXiv preprint arXiv:1805.12219}, 2018.

\bibitem{yu2017deep}
X.~Yu, X.~Wu, C.~Luo, and P.~Ren, ``Deep learning in remote sensing scene
  classification: a data augmentation enhanced convolutional neural network
  framework,'' \emph{GIScience \& Remote Sensing}, vol.~54, no.~5, pp.
  741--758, 2017.

\bibitem{hu2014unsupervised}
H.~Hu and Y.~Ban, ``Unsupervised change detection in multitemporal sar images
  over large urban areas,'' \emph{IEEE Journal of Selected Topics in Applied
  Earth Observations and Remote Sensing}, vol.~7, no.~8, pp. 3248--3261, 2014.

\bibitem{lee1981speckle}
J.-S. Lee, ``Speckle analysis and smoothing of synthetic aperture radar
  images,'' \emph{Computer graphics and image processing}, vol.~17, no.~1, pp.
  24--32, 1981.

\bibitem{otsu1979threshold}
N.~Otsu, ``A threshold selection method from gray-level histograms,''
  \emph{IEEE transactions on systems, man, and cybernetics}, vol.~9, no.~1, pp.
  62--66, 1979.

\bibitem{yen1995new}
J.-C. Yen, F.-J. Chang, and S.~Chang, ``A new criterion for automatic
  multilevel thresholding,'' \emph{IEEE Transactions on Image Processing},
  vol.~4, no.~3, pp. 370--378, 1995.

\bibitem{MJ_Flourens}
``Cmarie-jean-pierre flourens,''
  \url{https://www.britannica.com/biography/Marie-Jean-Pierre-Flourens},
  accessed: 2020-04-11.

\end{thebibliography}




\end{document}